%% file: MoM-GAN.tex
\documentclass{article}
\usepackage[utf8]{inputenc}
\usepackage{fullpage}
\usepackage{amsfonts}
\usepackage{amsmath}
\usepackage{amssymb}
\usepackage{graphicx}
\usepackage{accents}
\usepackage{import}
\usepackage{bbm}
\usepackage{enumerate}

\usepackage[square]{natbib}
\usepackage{bm}
\usepackage{hyperref}
\hypersetup{
	colorlinks=true,
	linkcolor=blue,
	filecolor=blue,
	citecolor = black,      
	urlcolor=blue,
} % set the link colors
\usepackage{subfig}
\usepackage{color}
\usepackage[ruled,linesnumbered]{algorithm2e}
\usepackage{caption}

\usepackage{booktabs}
\input{myMacros}

\graphicspath{{figures/}}

\numberwithin{equation}{section}
\theoremstyle{plain}
\newtheorem{theorem}{Theorem}[section]

\newtheorem{lemma}{Lemma}[section]

\newtheorem{proposition}{Proposition}[section]
\newtheorem{corollary}{Corollary}[section]

\newcommand{\PP}{\mathbb{P}}
\newcommand{\Var}{\mathbb{V}\mathrm{ar}}
\newcommand{\E}{\mathbb{E}}
\newcommand{\composition}{\circ}
\newcommand{\R}{\mathbb{R}}

\newcommand{\data}{\boldsymbol{x}}
\newcommand{\samplesize}{n}
\newcommand{\samplesizeNoise}{m}
\newcommand{\noise}{\boldsymbol{z}}
\newcommand{\dimNoise}{k}
\newcommand{\dimInput}{p}

\newcommand{\distriT}{\mu}
\newcommand{\distriN}{\nu} % supposed distribution to generate noise
\newcommand{\generator}{g}
\newcommand{\genT}{\generator^*}
\newcommand{\genEst}{\widehat{\generator}}
\newcommand{\genSpace}{\mathcal{G}}
\newcommand{\discriminator}{d}

\newcommand{\disSpace}{\mathcal{D}}
\newcommand{\distri}{\gamma}
\newcommand{\domain}{\Omega}
\newcommand{\distriTemp}{\widehat\distriT_\samplesize}
\newcommand{\domainN}{\Gamma}
\newcommand{\sdDis}{\sigma}
\newcommand{\varianceDis}{\sdDis^2}
\newcommand{\varRade}{\xi}

\newcommand{\Rade}{\mathcal{R}}
\newcommand{\RadeN}{\Rade_{\samplesize}}

\newcommand{\momgan}{\mathrm{MG}}
\newcommand{\MOM}{\mathrm{MoM}}
\newcommand{\blockNum}{K}

\newcommand{\blockIndx}{B}
\newcommand{\quantile}{Q}

\newcommand{\nn}{\mathcal{NN}}
\newcommand{\Width}{N}
\newcommand{\Depth}{L}
\newcommand{\disWidth}{\Width_{\disSpace}}
\newcommand{\disDepth}{\Depth_{\disSpace}}
\newcommand{\genWidth}{\Width_{\genSpace}}
\newcommand{\genDepth}{\Depth_{\genSpace}}

\newcommand{\approxErr}{\mathcal{E}}

\newcommand{\fcn}{f}
\newcommand{\fcnEva}{h}
\newcommand{\distance}{d}
\newcommand{\evaSpace}{\mathcal{H}}
\newcommand{\lipC}{c}
\newcommand{\lipSpace}{\mathrm{Lip}}
\newcommand{\lipSupB}{B}

\newcommand{\norm}[1]{\ensuremath{|\!| #1 |\!|_2}}

\newcommand{\normsup}[1]{\ensuremath{\hspace{0.5mm}\!|\!| #1 | \! |_{\infty}}}

\UseRawInputEncoding

\title{Distribution Estimation of Contaminated Data via DNN-based MoM-GANs}
\date{}

\begin{document}
	\author{Fang Xie,~Lihu Xu,~Qiuran Yao~and Huiming Zhang
%		\thanks{F. Xie is supported in part by UIC Research Grant with No. of UICR0700022-22 at BNU-HKBU United International College, Zhuhai, PR China. 
%			L. Xu is supported in part by NSFC grant (No.12071499), Macao S.A.R grant FDCT  0090/2019/A2 and University of Macau grant  MYRG2018-00133-FST. 
%			H. Zhang is supported in part by NSFC grant (No.12101630) and the University of Macau under UM Macao Talent Programme (UMMTP-2020-01).}		
%		\thanks{Fang Xie is with Division of Science and Technology, BNU-HKBU United International College,
%			Zhuhai, China (e-mail: fangxie@uic.edu.cn).}
%		%
%		\thanks{Lihu Xu and Qiuran Yao are with 1. Department of Mathematics, Faculty of Science and Technology, University of Macau, Av. Padre Tom\'{a}s Pereira, Taipa Macau, China; 2. Zhuhai UM Science \& Technology Research Institute, Zhuhai, China (e-mail: lihuxu@um.edu.mo, yb97478@connect.um.edu.mo).}
%		%
%		\thanks{Huiming Zhang is with  1. Institute of Artificial Intelligence, Beihang University, Beijing, 100083, China; 2. Zhuhai UM Science \& Technology Research Institute, Zhuhai, China; 3. Department of Mathematics, Faculty of Science and Technology, University of Macau, Av. Padre Tom\'{a}s Pereira, Taipa Macau, China; (e-mail: zhanghuiming@buaa.edu.cn~(Corresponding author)).}	
\thanks{\textit{Corresponding author (e-mail: zhanghuiming@buaa.edu.cn)}}
}
		
\maketitle
	
	\begin{abstract}
		This paper studies the distribution estimation of contaminated data by the MoM-GAN method, which combines generative adversarial net (GAN) and median-of-mean (MoM) estimation. 
		We use a deep neural network (DNN) with a ReLU activation function to model the generator and discriminator of the GAN. 
		Theoretically, we derive a non-asymptotic error bound for the DNN-based MoM-GAN estimator measured by integral probability metrics with the $b$-smoothness H\"{o}lder class.
		The error bound decreases essentially as $\samplesize^{-b/\dimInput}\vee n^{-1 / 2}$, where $\samplesize$ and $\dimInput$ are the sample size and the dimension of input data. 
		We give an algorithm for the MoM-GAN method and implement it through two real applications.
		The numerical results show that the MoM-GAN outperforms other competitive methods when dealing with contaminated data.
		
		
	\end{abstract}

\section{Introduction}\label{sec:intro}
%\subsection{Backgrounds}
Generative adversarial network (GAN) \citep{Goodfellow2014Generative} as one of the dominant deep learning methods has shown its success in academia and industries.
It can estimate the generative models and hence determine whether a sample is from the model distribution or data distribution via adversarial nets.
The generative and discriminative models are usually implemented by neural networks, such as ReLU neural networks.
To train a neural network well, it requires a large amount of data.
Nevertheless, in the era of big data, data deluge has been a conundrum, and the consequent data contamination in data collection processes is inevitable.
When encountering contaminated data, inaccurate neural networks will be produced with low accuracy in the training processes.
In other words, the trained neural networks may be very sensitive to outliers, which are probably not significant to the real problems.
To solve this problem, robust estimation, as an important statistical method, should also be considered in deep learning to stabilize the models/methods.

With its wonderful robustness property, the median-of-mean (MoM) estimator has been well-studied in statistics.
Recently, the MoM approach has attracted more and more attention in the robust deep-learning area.
\citet{Staerman2021OT} combined the MoM estimation and the GAN framework to propose a robust learning method, which is called \emph{MoM-GAN} or MoM-WGAN if the discrimination is measured by a Wasserstein distance. This learning method has the merits of MoM and GAN and is more stable than traditional GANs. In this paper, we model the generator and discriminator in the GAN by deep neural networks (DNNs) and establish the theory of the DNN-based MoM-GAN method rigorously. More precisely, we derive the convergence rate of the proposed estimator measured by integral probability metric under the H\"{o}lder evaluation function class, which includes the Wasserstein-1 distance. 
Other robust methods related to the MoM approach can refer to Section~\ref{sec:relatedworks}.

\subsection{Contributions}
In summary, our main contributions are given as follows:
\begin{itemize}
	\item We use the DNN with ReLU activation function for both the discriminator and generator of GAN. In the setting of contaminated data, we establish non-asymptotic error bounds of our DNN-based MoM-GANs measured by integral probability metrics with the H\"{o}lder function class.
	The error bound (in Theorem~\ref{thm:main}) shows an explicit relation between error and parameters (e.g., width and depth of DNN, the sample size of data).
	
	\item For data assumption, our DNN-based MoM-GANs do not depend on any assumption on the distribution of the outliers; thus it is robust to adversarial contamination. We derive finite sample high probability error bounds for the associated MoM estimator, which depend on the number of the outliers and sane blocks, see Propositions \ref{lem:our} and \ref{prop:genApproxErr}.
	
	\item We propose an algorithm to implement the DNN-based MoM-GANs; the experiments show that our method can effectively overcome noise disturbance and has stronger robustness than Wasserstein GAN. Furthermore, we use binary search to find the optimal block numbers $K$.
\end{itemize}

\subsection{Related Works}\label{sec:relatedworks}
% \subsubsection{GAN-based Methods}\label{subsec:RW-GAN}
\paragraph{GAN-based Methods} Assume $\data_1,\dots,\data_\samplesize\in\domain\subset\R^\dimInput$ are independent observations from the distribution $\distriT$. Suppose $\distriN$ is the distribution on $\domainN\subset\R^\dimNoise$ to generate the noise/fake signal.
Let $\generator\in\genSpace$ be a generator function from $\domainN$ to $\domain$.
For simplicity, $\R^\dimNoise$ can be assumed to be $\R$ with $\dimNoise=1$ \citep[Remark~2.3]{huang2021error}.
Let $\discriminator\in\disSpace$ be a discriminator function from $\R^\dimInput$ to $\R$.
In general, the two function spaces $\genSpace$ and $\disSpace$ are supposed to be the collections of neural networks, for example, those with a certain width and depth collections.
%Denote the composition of two functions $\discriminator$ and $\generator$ by $\discriminator\composition\generator$.

The generative adversarial network for learning the data distribution $\distriT$ by training a generator $\generator$ and a discriminator $\discriminator$ is to solve the following min-max problems. The oracle GAN generator $\genT$ and its empirical estimator $\genEst$ can be defined as
\begin{align}\label{eq:oracle}
	\genT&\in\arg\min_{\generator\in\genSpace}\max_{\discriminator\in\disSpace}\{\E_{\data\sim\distriT}[\discriminator(\data)]-\E_{\noise\sim\distriN}[\discriminator(\generator(\noise))]\};
	\nonumber\\
	\genEst&\in\arg\min_{\generator\in\genSpace}\max_{\discriminator\in\disSpace}\{\frac{1}{n} \sum_{i=1}^{n} \discriminator(\data_i)-\frac{1}{\samplesizeNoise}\sum_{i=1}^{\samplesizeNoise}\discriminator(\generator(\noise_i))\}.
\end{align}
For matching the reality, here we use the empirical average to replace the second term in \eqref{eq:oracle} based on i.i.d. sample $(\noise_{1},\dots,\noise_{m})$ drawn from $\distriN$. The GAN method can be considered as minimizing the dual form of Wasserstein-1 distance \citep{Kantorovich1958Space} between distributions $\distriN$ and $\distriT$ over the generator space.

With the success of the GAN framework, many variations of GANs have been developed to extend the boundaries of GANs.
A lot of literature studied whether the GANs can produce excellent estimation.
\citet{liang2021well,huang2021error} studied the convergence rates of GAN estimators under the integral probability metric. \citep{zhu2020deconstructing} studied a fundamental trade-off in the approximation error and statistical error in GANs. 

% \subsubsection{Robust Deep Learning Methods}\label{subsec:RW-MoM}
\paragraph{Robust Deep Learning Methods}
With collecting data automatically through a machine/algorithm, low-quality data with contamination are often produced.
When training models based on these data, misleading results consequently arise if using the traditional deep learning methods. \citet{Arora2017Generalization} showed that training of GAN might not own good generalization properties, such as perfect training but a badly trained distribution that is far away from the target distribution.
To reduce the destabilization caused by the contaminative data, robust deep learning methods have attracted more and more attention in recent years. To stabilize GANs, \citet{liu2021robust} proposed Wasserstein GAN-based estimation under Huber's contamination model, but their estimator is not robust to adversarial contamination.
\citet{Wang2019Direct} showed the robustness of their reformative adversarial network approach through several experiments.
As one of the successful robust estimators in statistics, the MoM estimator has also brought a trend in robust deep learning in recent years.  
\citet{Staerman2021OT} introduced outlier-robust estimators of the Wasserstein distance based on MoM methodology. They proved the strong consistency of the distribution estimators under the supposed MoM measure and applied their method to the Wasserstein GAN.
They only showed numerical results for the MoM-based WGAN but no theoretical guarantees in terms of DNN. 
\citet{Huang2021Deepmom} proposed a robust deep learning method called DeepMoM, which combined the MoM approach and Le Cam's principle \cite{Lecue2020Robust}.
They provided a simple gradient-based algorithm for their approach and studied through the simulated data of different levels of corruption.
Based on Lipschitz and convex loss functions, \citet{Chinot2020Robust} studied estimation and excess risk bounds for empirical risk minimizers and minimax MoM estimators.
Except for the MoM estimator, Huber M-estimator and least absolute deviation estimator are robust alternative estimators.
\citet{Lederer2020Risk} established empirical-risk bound for robust deep learning with Lipschitz-continuous loss functions, which included Huber loss and $\ell_1$ loss. 
\citet{Huang2021Deepmom} proposed a robust deep learning method called DeepMoM, which combined the MoM approach and Le Cam's principle \citep{Lecue2020Robust}.
They provided a simple gradient-based algorithm for their method and studied the simulated data with corruption at different levels. 
Recently, \citet{xu2022non} studied the excess risk bound of robust deep neural network regressions by the proposed log-truncated M-estimator under the condition that the data have $(1+\varepsilon)$-th moment with $\varepsilon \in (0,1]$.

The remaining paper is organized as follows. 
The preliminaries and notations are introduced in Section~\ref{subsec:notations}. 
The method and main results are introduced in Sections~\ref{sec:method} and~\ref{sec:theory}.
We give an algorithm and conduct two real applications in Section~\ref{sec:simApp}.
A conclusion and future work are given in Section~\ref{sec:conclusion}.
We defer the proofs to Section~\ref{sec:proofs} and Appendix.

\subsection{Preliminaries and Notations}\label{subsec:notations}
Denote $[\samplesize]:=\{1,\dots,\samplesize\}$.  Let $\nu$ be a measure on $\mathbb{R}^{m}$ and $\phi: \mathbb{R}^{m} \rightarrow \mathbb{R}^{\dimInput}$ a measurable mapping, where $m, \dimInput \in \mathbb{N}$. The push-forward measure $\phi_{\#} \nu$ of a measurable set $A \subseteq \mathbb{R}^{\dimInput}$ is defined as $\phi_{\#} \nu(A):=\nu\left(\phi^{-1}(A)\right)$. Let $\|A\|_{F}:=\sqrt{\sum_{i = 1}^m \sum_{j = 1}^n a_{ij}^2}$ be the Frobenius norm of a square matrix $A =(a_{ij}) \in \mathbb{R}^{m \times n}$.
For a real-valued function $f:\mathcal{X}\rightarrow \R$, we define $\|f\|_\infty:=\sup_{\bm{x}\in\mathcal{X}}|f(\bm{x})|$. Let $\lipSpace(\domain, \lipC)$ be all the real-valued Lipschitz functions from $\domain$ to $\R$ with Lipschitz constant $\lipC$, that is, 
\begin{align*}
	\lipSpace(\domain, \lipC):=\bigl\{\fcn:\domain\rightarrow\R: |\fcn({\bm z}_1)-\fcn({\bm z}_2)|\le c\norm{{\bm z}_1-{\bm z}_2}{\rm ~for\ any\ } {\bm z}_1,{\bm z}_2\in\domain\bigr\}.
\end{align*}
For any $B>0$, denote $$\lipSpace(\domain,\lipC,\lipSupB):=\{\fcn\in\lipSpace(\domain,\lipC):\normsup{\fcn}\le\lipSupB\}.$$ 

For $b>0$ with $b=s+q$, where $s \in \mathbb{N}_{0}:=\mathbb{N} \cup\{0\}, q \in(0,1], \dimInput \in \mathbb{N}$, denote H{\"o}lder class with smoothness index $b$  as
\begin{align*}
	\mathcal{H}^{b}(\mathbb{R}^{p}):=\{f: \mathbb{R}^{p} \rightarrow \mathbb{R}, \max\limits_{\|\alpha\|_{1} \leq s}\left\|\partial^{\alpha} f\right\|_{\infty} \leq 1, \max\limits_{\|\alpha\|_{1}=s} \sup\limits_{x \neq y} \frac{\left|\partial^{\alpha} f(x)-\partial^{\alpha} f(x)\right|}{\|x-y\|_{2}^{q}}\leq 1\}.
\end{align*}
For any subset $\mathcal{X} \subseteq \mathbb{R}^{\dimInput}$, we denote $\mathcal{H}^{b}(\mathcal{X}):=\{f: \mathcal{X} \rightarrow \mathbb{R}, f \in \mathcal{H}^{b}(\mathbb{R}^{\dimInput})\}$.

Let $\data_1,\dots,\data_\samplesize\in\R^\dimInput$ be independent samples from some distribution. Let $\varRade_1,\dots,\varRade_\samplesize$ be independent Rademacher random variables. Let $\mathcal{F}$ be a class of functions $f:\R^\dimInput\mapsto\R$. The empirical Rademacher complexity is defined as
\begin{equation*}
	\RadeN(\mathcal{F};\data):=\E\sup_{f\in\mathcal{F}}\Bigl|\frac{1}{\samplesize}\sum_{i=1}^\samplesize\varRade_i f(\data_i)\Bigr|.
\end{equation*}

Let $\nu$ and $\gamma$ be a pair of probability measure on a space $\mathcal{X}$, and let $\mathcal{F}$ be a class of functions $f:\mathcal{X} \rightarrow \mathbb{R}$ that are integrable w.r.t. $\nu$ and $\gamma$. Define the \emph{Integral Probability Metric} (IPM, see \cite{muller1997integral}) as
\begin{align*}
	d_{\mathcal{F}}(\nu, \gamma):=\sup _{f \in \mathcal{F}}\left|\int f(d \nu-d \gamma)\right|=\sup _{f \in \mathcal{F}}\left|\mathbb{E}_{\nu}[f(X)]-\mathbb{E}_{\gamma}[f(Z)]\right|,
\end{align*}
which is a pseudo-metric (not necessarily distinguish points). When $\mathcal{F}$ is sufficiently rich, then $d_{\mathcal{F}}$ becomes a metric, such as Kolmogorov metric and Total Variation (TV) distance. IPM is well-studied in probability theory, mass transportation problems, etc. It can be used to study GANs. Various GANs can be obtained by choosing specific $\mathcal{F}$:
\begin{enumerate}
	\item $\mathcal{F}=\operatorname{Lip}\left(\mathbb{R}^{p}, 1, 1\right)=\mathcal{H}^{1}(\mathbb{R}^{p})$ is the bounded 1-Lipschitz function class, then $d_{\mathcal{F}}$ is the \emph{Wasserstein-1 distance},
	\begin{center}
		$\mathcal{W}_{1}(\mu, \nu):=\inf _{\gamma \in \Gamma(\mu, \nu)} \int_{\mathcal{X} \times \mathcal{X}} {\|x-y\|_{2}} \mathrm{~d} \gamma(x, y)$,
	\end{center}
	where $\Gamma(\mu, \nu)$ is the set of all couplings of $(\mu, \nu)$. $\mathcal{W}_{1}(\cdot, \cdot)$ leads to Wasserstein GAN \citep{arjovsky2017wasserstein};

	\item $\mathcal{F}=\mathcal{H}^{0}(\mathbb{R}^{p}):=\left\{f \mid\|f\|_{\infty} \leq 1\right\}$, then $d_{\mathcal{F}}$ is the TV distance. It gives TV GAN \citep{chao2018robust};
	
	\item $\mathcal{F}$ is a class of functions with bounded RKHS norm, then $d_{\mathcal{F}}$ is the \emph{maximum mean discrepancy}. \citet{li2015generative} studied the corresponding  generative moment matching network (GMMN) that differs from GANs.
\end{enumerate}

To serve our main theory, we define the DNNs below. For the functional class of discriminator and generator, we consider the feed-forward DNN with $\Depth$ hidden layers as the $\Depth$-compositional function class indexed by the parameter $\bm\beta$: 
\begin{align*}
    f (\data; \bm\beta):& \mathbb{R}^{\dimInput} \to \mathbb{R}\\
    &\data\mapsto W_{\Depth} \sigma_{\Depth}\left(\cdots W_{1}\sigma_{1}\left(W_{0} {\data}\right)\right),\ \ \  \bm\beta: = (W_0, \ldots, W_\Depth),
\end{align*}
where $W_{l} \in \mathbb{R}^{D_{l+1} \times D_{l}}$ for $l=0,1,2,\dots,\Depth$ with $D_l\in\mathbb{N}$,  $D_{L+1}=1$, and $D_0=\dimInput$, activation functions $\{\sigma_j\}_{j=1}^\Depth: \mathbb{R}^{D_j} \to \mathbb{R}^{D_j}$ are multivariate mappings with each component.

We define the \textit{width} and \textit{depth} of the DNN by $N=\max \left\{D_{1}, \ldots, D_{\Depth}\right\}$ and $L$ respectively. The \textit{size} of the DNN is the number of parameters and defined as $S=\sum_{i=0}^{\Depth}D_{i+1} \times D_{i}\sim \Depth\Width^2$.

The collection of DNNs is denoted by
\begin{align}\label{eq:NN}
	\nn(\Width, \Depth) := \big\{ f ({\data}; \bm\beta) =W_{\Depth} \sigma_{\Depth}\left(\cdots W_{1}\sigma_{1}\left(W_{0} {\data}\right)\right) |\, \bm\beta: = (W_0, \ldots, W_\Depth )\big\}.
\end{align}
In below, we denote the set of functions by $\nn(\Width, \Depth)$  that neural networks can represent with width at most $\Width$ and depth at most $\Depth$ in \eqref{eq:NN}.

\section{MoM-GAN Method}\label{sec:method}
This section will introduce the robust MoM-GAN method, which is used to construct an estimator for unknown probability measures, and we first present the outliers and DNN model assumptions.

\subsection{Data and DNN Model Assumptions}

In practical applications, we often encounter adversarial contamination of the data set, which means some of the data are
replaced by other data. The replaced data contains outliers that can make the regular method work badly. For example, consider the $O \cup I$ frameworks:

\textbf{A.1} \textit{Suppose that the data $\data_1,\dots,\data_\samplesize$ contains $n-n_o$ inliers $\{\data_{i}\}_{i \in I}$ drawn i.i.d. from the target distribution $\distriT$ that has support $[0,1]^{\dimInput}$, and there are no distributional assumptions on $n_o$ outliers $\{\data_{i}\}_{i \in O}$. We suppose that the dimension $\dimInput\in\mathbb{N}$ is a fixed number.}

Assuming $\dimInput$ is a finite constant is reasonable since if we consider $\data_i$ is the vectorization of a picture, then $\dimInput$ is the number of pixels that is finite. Under A.1, rather than taking the average of all the observations, the median-of-mean (MoM) estimator was introduced by \citet{nemirovskij1983problem}. The sample is equally split into several non-overlapping blocks, and the MoM estimator is then defined as the
median of each empirical mean for the block sample. The assumption of the MoM estimator is more relaxed than the assumption of classical Huber's contamination model, see Section 1.2 in \citet{Lerasle2019Lecture}. MoM estimator eliminates any assumption on the outliers by the following two assumptions, while Huber's contamination model requires i.i.d. assumption for $\{\data_{i}\}_{i \in [n]}$ and the estimation error increases dramatically
when the contamination ratio increases; see \cite{chen2016general}.

Formally, we consider $\blockNum\in [n]$ blocks $\blockIndx_1,\dots,\blockIndx_\blockNum\subset [n]$ such that $\blockIndx_j\cap\blockIndx_k=\emptyset$ for any $1\le j<k\le \blockNum$ and $\cup_{k=1}^{\blockNum}\blockIndx_k=[n]$.
For simplicity, we assume that $\blockNum$ is a factor of $\samplesize$ and the $\samplesize$ samples can be divided into $\blockNum$ blocks with equal cardinality (each of the blocks contains $\samplesize/\blockNum$ samples). We remove some observations if $n$ cannot be divided by $K$.
In each block, given any function $\fcn:\R^\dimInput\mapsto\R$, we define its empirical mean by
\begin{equation*}
	\E_{\blockIndx_k}[\fcn]:=\frac{1}{|\blockIndx_k|}\sum_{i\in\blockIndx_k}\fcn(\data_i)=\frac{\blockNum}{\samplesize}\sum_{i\in\blockIndx_k}\fcn(\data_i),~\forall~k\in\{1,\dots,\blockNum\},
\end{equation*}
where $|\blockIndx_i|$ denotes the cardinality of the set $\blockIndx_i$. 

For any $\alpha\in(0,1)$, we denote $\quantile_\alpha[\fcn]$ the $\alpha$-quantile of the data $\{\E_{\blockIndx_1}[\fcn],\dots,\E_{\blockIndx_K}[\fcn]\}$.
The MoM of the function~$\fcn$ among $\blockNum$ blocks based on the samples $\data_1,\dots,\data_\samplesize$ is defined by
\begin{equation} \label{eq:med}
	\MOM_{\blockNum}[\fcn]:=\max_{k\in\{1,\dots,\blockNum\}}\bigl\{\E_{\blockIndx_k}[f]: \E_{\blockIndx_k}[f] \le\quantile_{1/2}[\fcn]\bigr\}.
\end{equation}

To serve for error bounds in the presence of outliers, we require specific outlier assumptions. Consider the specific fraction function of the polluted inputs in \citet{laforgue2021generalization}:\\

\textbf{A.2}  \textit{$K$ can be written as $K=K_O+K_S$, where $K_O$ represents the number of blocks containing at least one outlier and $K_S$ represents the number of the sane block containing no outliers. Let $\varepsilon:=n_o/n$ be the fraction of the outliers in the data. There exists a fraction function $\eta(\varepsilon)$ for sane block such that
	\begin{equation}\label{eq:eta}
		K_S \ge \eta(\varepsilon) K~\text{for a function}~ \eta(\varepsilon)>0.
	\end{equation}
	Without loss of generality, we assume that $\{\data_{i}\}_{i \in B_1}$ has no outliers.}

\begin{remark}
	For example, under A.1 and A.2, our constraint for the fraction of the polluted inputs is
	$n_o=\varepsilon n<\frac{K}{2}\le\frac{n}{2}.$  Define a function of the outlier fraction $\alpha(\varepsilon):[0,0.5]\rightarrow [0,1]$ such that
	$2\varepsilon<\alpha(\varepsilon)<1$ and $\alpha(\varepsilon)n\le K$.
	It follows that
	$K_S=K-K_O\ge K-n_o=K-\varepsilon n \ge [1-\varepsilon/\alpha(\varepsilon)]K.$
	Thus we could choose $\eta(\varepsilon)=1-\varepsilon/\alpha(\varepsilon)$ in \eqref{eq:eta}.
\end{remark}

For DNN, we need the following assumption.\\

\textbf{A.3} \textit{ Suppose that the parameter set for \eqref{eq:NN} has $F$-norm ball with radius $r$ and  maximum spectral norm bounded by $M$,
	\begin{equation*}\label{eq:FBALL}
		\bm\Theta:=\{\bm\theta:=( W_0, \ldots, W_{L} ):\|\bm\theta\|_{\mathrm{F}} \le r~\text{and}~\max_{0 \le l \le L}{\lambda _{\max }}({W_l}) \le {M}\} \subseteq \mathbb{R}^{S}.
	\end{equation*}
}

\subsection{A Robust GAN Estimator Based on Median-of-mean}
To make GAN method adapt to the corrupted data, we recommend a method by replacing the empirical mean of the discriminator in~\eqref{eq:oracle} with the empirical median-of-mean of the discriminator on the samples ${\bm{X}}:=(\data_{1},\dots,\data_{n})$ with outliers. The \emph{MoM-GAN} estimator $\genEst_{\momgan}$ is defined as
\begin{equation}\label{def:momgan}
	\genEst_{\momgan}\in\arg\min_{\generator\in\genSpace}\max_{\discriminator\in\disSpace}\Big\{\MOM_{\blockNum}[\discriminator]-\frac{1}{\samplesizeNoise}\sum_{i=1}^{\samplesizeNoise}\discriminator(\generator(\noise_i))\Big\},
\end{equation}
where $\MOM_{\blockNum}[\discriminator]$ is the median-of-mean of the discriminator among $\blockNum$ blocks based on the sample ${\bm{X}}$ and defined in~\eqref{eq:med}. The push-forward measure based on MoM-GAN estimator $(\widehat\generator_{\momgan})_{\#}\distriN$  will be used to estimate the unknown probability measure $\distriT$.

%For notation simplicity, we use $\MOM_\blockNum[\discriminator]$ instead of $\MOM_{\blockNum}[\discriminator|\data_1,\dots,\data_\samplesize]$.

% If the expectation term in \eqref{def:momgan} is replaced by its empirical form, then the estimator is the same as the MoM-based WGAN estimator that was defined in Section 4.3 of \cite{Staerman2021OT}.

For \eqref{def:momgan}, \citet{Staerman2021OT} only showed the numerical results for this method but no theoretical guarantees.
In this paper, we develop the theory of MoM-GANs based on DNNs approximation theory and the concentration inequality of suprema of MoM empirical process.
%In fact, \citet{Staerman2021OT} formulated the MoM-GAN estimators with real data analysis and no theoretical results is investigated.
We construct neural network functions for generator and discriminator with a certain width, depth, and Lipschitz conditions, and non-asymptotic error bound on the MoM-GAN estimator is established in Theorem \ref{thm:main} below. \\

\section{Main Results}\label{sec:theory}

In this section, we use IPM to evaluate the performance of our MoM-GAN estimator.

Recall that $\generator_{\#}\distriN$ is the push-forward distribution under $\generator$.  Given a real-valued function $\fcnEva\in\evaSpace:=\evaSpace^b(\R^\dimInput)$, we define the following IPM
\begin{align*}
	\distance_{\evaSpace}(\distriT,(\genEst_{\momgan})_{\#}\distriN):=\sup_{\fcnEva\in\evaSpace}\Bigl\{\E_{\data\sim\distriT}[\fcnEva(\data)]-\E_{\noise\sim\distriN}[\fcnEva(\genEst_{\momgan}(\noise))]\Bigr\}
\end{align*}
to evaluate the distance between the target $\distriT$ and the estimated distribution $(\genEst_{\momgan})_{\#}\distriN$ under the \emph{evaluation class} $\evaSpace$.

In our theory, we consider the neural networks $\nn(\Width,\Depth)$ in \eqref{eq:NN} with ReLU activation functions $\{\sigma_j(\bm{z})=(z_1 \vee 0,\cdots,z_{D_j}\vee 0)$ for $\bm{z}:=(z_1,\ldots,z_{D_j})^\top\in \mathbb{R}^{D_j}\}_{j=1}^L$. For convenience, $\disSpace$ is assumed in \textit{symmetric class} (if $d \in \disSpace$, then $-d \in \disSpace$). Based on the notations and assumptions above, we present the main result.

\begin{theorem}[Non-asymptotic Bound]\label{thm:main}
% 	Assume target measure $\mu$ is absolutely continuous on $\mathbb{R}$ and $\generator_{\#}\distriN$ is supported on $[0,1]^{\dimInput}$ for all $\generator\in \mathcal{G}$. 
	Assume target measure $\mu$ is supported on $[0,1]^{\dimInput}$, and the distribution $\nu$ is absolutely continuous on $\mathbb{R}$. 
	Suppose $\sup_{\discriminator\in\disSpace}\Var(\discriminator(\data))=:\varianceDis<\infty$ and ${\E}\max_{i\in [\samplesizeNoise]}{\left\| \noise_i\right\|_2}<\infty$ with $\data\sim\mu$, $\noise_i\sim\nu$, $i\in[\samplesizeNoise]$.
	The evaluation class is $\mathcal{H}=\mathcal{H}^{b}\left([0,1]^{\dimInput}\right)$
	with ${b}=s+q, s \in \mathbb{N}_{0}$ and $q \in(0,1].$
	If assumptions A.1-A.3 are satisfied, there exist a generator
	$\mathcal{G}=\left\{g \in \nn(\genWidth,\genDepth): g(\mathbb{R}) \subseteq[0,1]^{p}\right\}$  
	and a discriminator $\disSpace=\nn(\disWidth,\disDepth) \cap \operatorname{Lip}\left(\mathbb{R}^{\dimInput}, J, 1\right)$ with $\genWidth\ge7\dimInput+1$, $\genDepth\ge2$,
	\begin{equation}\label{eq:sample}
		n \leq (1-\varepsilon)^{-1}\left(\frac{\genWidth-p-1}{2}\left\lfloor\frac{\genWidth-p-1}{6 p}\right\rfloor\left\lfloor\frac{\genDepth}{2}\right\rfloor+2\right)\wedge\left[\disWidth\disDepth / C_{p}\right]^{2},
	\end{equation}
	$C_\dimInput$ being a positive constant and
	$J \le (s+1) p^{s+1 / 2} L_{\disSpace}(\disWidth\disDepth)^{{\color{black}{((4 {b}-4)\vee 0)}} / p}(1260 \disWidth^{2}\disDepth^{2} 2^{L_{\disSpace}^{2}}+19 s 7^{s})$,
	s.t. MoM-GAN estimator $\genEst_{\momgan}$ satisfies
			\begin{align}
				\distance_{\evaSpace}(\distriT,(\genEst_{\momgan})_{\#}\distriN)&\le O(n^{-{b} / p})
				\label{eq:main1}\\
				& + O\left(\frac{1}{\sqrt{n}}{\Bigg[rM^{\disDepth}}\sqrt \disDepth+\sqrt{\frac{\blockNum}{(1-\varepsilon)}}\Bigg]\right)+O\left(\sqrt{\frac{K\disDepth}{n}}\frac{M^{\disDepth}}{[1-(2\eta(\varepsilon))^{-1}]}\right)\vee O\left(\frac{K}{\sqrt{[1-(2\eta(\varepsilon))^{-1}]n}}\right)\label{eq:main2}\\
				& +O\left(r\sqrt{\frac{\disDepth+\genDepth}{m}}{M^{{\disDepth+\genDepth}}}+\sqrt{\frac{t}{m}}\right)\label{eq:main3}
			\end{align}

	with probability at least $1-{\rm{e}}^{-\blockNum/32}-2{\rm{e}}^{-\frac{K}{8\eta(\varepsilon)}\left(1-\frac{1}{2\eta(\varepsilon)}\right)^2}-{\rm{e}}^{-t}$ for $t >0$.
\end{theorem}

%\FX{Update the main results for the empirical form.}

\begin{remark}
	The first term \eqref{eq:main1} is the bias term due to the discriminator approximation error, which has order $O(n^{-{b} /p})$. For fixed $K$ and $\disDepth$, the terms in~\eqref{eq:main2} have the order $O(n^{-1 / 2})$ , which is the variance term derived from: 1) the generator approximation error for large size of the generator network $\mathcal{G}$; 2) concentration for suprema of the empirical processes in term of MoM-estimator. The remainder term \eqref{eq:main3} is obtained by the concentration inequalities for suprema of the empirical process. If setting $\samplesizeNoise=\samplesize$, we can get the order $O(\samplesize^{-1/2})$ for term~\eqref{eq:main3}. So, the upper bound of $\distance_{\evaSpace}(\distriT,(\genEst_{\momgan})_{\#}\distriN)$ has the order $O(\samplesize^{-b/\dimInput}\vee\samplesize^{-1/2})$. In \eqref{eq:sample}, for a given $n$ and $p$, one requires that $\nn(\genWidth,\genWidth)$ and $\nn(\disWidth,\disWidth)$ satisfy the upper bound restriction $n \le O(\genWidth^2 \genDepth\wedge\left[\disWidth\disDepth\right]^{2})$ in order to balance the generator and the discriminator approximation errors. 
\end{remark}

For fixed $K$ and $\disDepth$, when $p>2b$, $n^{-{b} / p} \vee n^{-1 / 2}>n^{-1 / 2}$, it can been seen that Theorem \ref{thm:main} gives slow non-parametric convergence rate that is derived from the infinite-parameter nature of the GAN problems. The specific constants in the bound can be found in the proof for interested readers. In details, the high probability upper bound and the exact constants in Theorem \ref{thm:main} is obtained by a new error decomposition lemma~(Lemma~\ref{lem:errDecomposition}), Proposition ~\ref{lem:our} and Proposition~\ref{prop:genApproxErr} in below.

\begin{remark}
	Compared with \citet{huang2021error}, which established the convergence rates of WGAN by the expected error bounds, we study non-asymptotic upper bound by deriving Rademacher complexity for deep neural networks rather than using the covering number of DNNs. Moreover, Theorem 22 in \citet{huang2021error} requires that $\{\data_{i}\}_{i \in [n]}$ are i.i.d. with finite exponential moment, our result allows that the outliers $\{\data_{i}\}_{i \in O}$ may be non-identically distributed and arbitrary unbounded.
\end{remark}

For ease of notation, we use $\E_{\distri}[\fcn]$ instead of $\E_{\data\sim\distri}[\fcn(\data)]$ for any distribution $\distri$ and function $\fcn$. Define the  discriminator approximation error of $\evaSpace$ from $\disSpace$ on $\domain$ as:
\begin{equation}\label{def:approxError}
	\approxErr(\evaSpace,\disSpace,\domain):=\sup_{\fcnEva\in\evaSpace}\inf_{\discriminator\in\disSpace}\|\fcnEva-\discriminator\|_{L^\infty(\domain)}.
\end{equation}

\begin{lemma}[Error Decomposition]
	\label{lem:errDecomposition}
	Suppose that $\distriT$ and $\generator_{\#}\distriN$ are supported on $\domain\subset\R^{\dimInput}$ for all $\generator\in\genSpace$ and assume $\disSpace$ is a \textit{symmetric class}.
	Let $\genEst_{\momgan}$ be defined in~\eqref{def:momgan}.
	Then, for any function class $\evaSpace$ defined on $\domain$,
	\begin{align*}
		\distance_{\evaSpace}(\distriT,(\genEst_{\momgan})_{\#}\distriN)
		\le 2\approxErr(\evaSpace,\disSpace,\domain)+\sup_{\discriminator\in\disSpace}\bigl\{|\E_{\distriT}[\discriminator]-\MOM_{\blockNum}[\discriminator]|\bigr\}+\inf_{\generator\in\genSpace}\sup_{\discriminator\in\disSpace}\{\MOM_{\blockNum}[\discriminator]-\E_{\generator_{\#}\distriN}[\discriminator]\}+2\distance_{\disSpace \circ \genSpace}(\widehat{\nu}_{m}, \nu),
	\end{align*}
	where $\widehat{\nu}_{m}:=\frac{1}{m}\sum_{i=1}^m\delta_{\noise_i}$.
\end{lemma}
%\FX{Update the error decomposition for the empirical form.}

From this lemma, we know that the evaluation error $\distance_{\evaSpace}(\distriT,(\genEst_{\momgan})_{\#}\distriN)$ is controlled by four terms.

1) $\approxErr(\evaSpace,\disSpace,\domain)$ is the approximation error of evaluation class $\evaSpace$ from discriminator space $\disSpace$ on $\domain$, which has well studied in Section 2.2.2 of \cite{huang2021error}.

2) $\sup_{\discriminator\in\disSpace}\bigl\{\E_{\distriT}[\discriminator]-\MOM_{\blockNum}[\discriminator]\bigr\}$ is the statistical error of the MoM estimator, which can be controlled by the concentration for suprema of the MoM estimator (see Proposition~\ref{lem:our} below).

3) $\inf_{\generator\in\genSpace}\sup_{\discriminator\in\disSpace}\{\MOM_{\blockNum}[\discriminator]-\E_{\generator_{\#}\distriN}[\discriminator]\}$ is the generator approximation
error, whose high-probability error bound can be obtained by Proposition~\ref{prop:genApproxErr} below.

4) $\distance_{\disSpace \circ \genSpace}(\widehat{\nu}_{m}, \nu)$ is the statistical error for the simulated samples in terms of IPM. Under the assumption $\sup_{\discriminator\in\disSpace}\normsup{\discriminator}\le\lipSupB$, the statistical error can be bounded with high probability, by using Massart's concentration inequality for suprema of the bounded empirical processes [see Theorem~9 in~\cite{massart2000constants}].

\begin{proposition}[Concentration for Supremum of MoM Empirical Processes]\label{lem:our}
	Under assumptions A.1 and A.2, we have
	$$P\left\{\sup_{f \in \mathcal{F}} \bigl|\MOM_\blockNum[f]-\E_{\distriT}[f]\bigr|\le \frac{16\Rade_{n/K}(\mathcal{F}; \data) }{1-(2\eta(\varepsilon))^{-1}} \vee \sqrt{\frac{4e^2U_{1n}(e) }{1-(2\eta(\varepsilon))^{-1}}}\right\}\ge 1-{\rm{e}}^{-\frac{K\eta(\varepsilon)}{8}\left(1-\frac{1}{2\eta(\varepsilon)}\right)^2},~~e> 0$$
	where the upper bound $U_{1n}(e)$ satisfies
	$K_S\sup_{f \in \mathcal{F}}P{\left\{|(\E_{B_1}-\E)f|\ge e/2\right\}}\le U_{1n}(e)$ for $e \ge \frac{16\Rade_{n/K}(\mathcal{F}; \data) }{1-(2\eta(\varepsilon))^{-1}} \vee \sqrt{\frac{4e^2U_{1n}(e) }{1-(2\eta(\varepsilon))^{-1}}}$.
\end{proposition}

The proof Lemma~\ref{lem:our} is different from Proposition~2 in \citet{laforgue2021generalization}, while our upper bound results has sharper constants than the method of Theorem~37 in~\citet{Lerasle2019Lecture}.

\begin{proposition}[Non-asymptotic Generator Approximation Error]\label{prop:genApproxErr}
	Suppose that  $\genSpace=\nn(\genWidth,\genDepth)$ satisfies $\genWidth\ge7\dimInput+1$, $\genDepth\ge2$ and
	$(1-\varepsilon)n\leq \frac{\genWidth-p-1}{2}\left\lfloor\frac{\genWidth-p-1}{6 p}\right\rfloor\left\lfloor\frac{\genDepth}{2}\right\rfloor+2 .$ Assume that $\sup_{\discriminator\in\disSpace}\normsup{\discriminator}\le\lipSupB$.
	Let $\distriN$ be an absolutely continuous probability distribution on $\R$. Suppose that $\distriT$ and $\generator_{\#}\distriN$ are supported on $\domain\subset\R^{\dimInput}$ for all $\generator\in\genSpace$. If $\sup_{\discriminator\in\disSpace}\Var(\discriminator(\data))=:\varianceDis<\infty$, and A.1 and A.2 are satisfied. Then there exists a $\generator\in\genSpace$ such that
	\begin{align*}
		\inf_{\generator\in\genSpace}\sup_{\discriminator\in\disSpace}\bigl\{\MOM_\blockNum[\discriminator]-\E_{\generator_{\#}\distriN}[\discriminator]\bigr\}\le 2\Rade_{(1-\varepsilon) n}(\disSpace; \data)+\frac{16\Rade_{n/K}(\disSpace; \data) }{1-(2\eta(\varepsilon))^{-1}}\vee \sqrt{ \frac{[(16\varianceDis)\wedge(32B^2/{\rm{e}})] K^2}{[1-(2\eta(\varepsilon))^{-1}]n}}+\lipSupB\sqrt{\frac{\blockNum}{(1-\varepsilon)n}}
	\end{align*}
	with probability at least $1-{\rm{e}}^{-\blockNum/32}-{\rm{e}}^{-\frac{K}{8\eta(\varepsilon)}\left(1-\frac{1}{2\eta(\varepsilon)}\right)^2}$.
\end{proposition}
%\FX{Update the probability bound above for the empirical form.}

To bound the Rademacher complexity in Proposition~\ref{prop:genApproxErr}, one needs the following result.

\begin{lemma}[Rademacher Complexity of Neural Networks]
	\label{lem:RadComplexityNN}
	Let $\{\data_i\}_{i = 1}^n$ be independent samples with finite ${\E}\max_{i \in [n]}{\left\| \data_i \right\|_2}$. For \eqref{eq:NN}, suppose that $\discriminator\in\nn(\Width,\Depth)$ whose activation functions are $\sigma_l: \mathbb{R}^{D_{l}} \rightarrow \mathbb{R}^{D_{l}}$ are $a$-Lipschitz with respect to Euclidean norm on their input and output spaces. If the parameter set has $F$-norm ball with radius $r$ and maximum spectral norm bounded by $M$, i.e.
	\begin{align*}\label{eq:FBALL}
		\bm\Theta:=\{\bm\theta&:=( W_0, \ldots, W_{\Depth} ):\|\bm\theta\|_{\mathrm{F}} \le r~\text{and}~\max_{0 \le l \le L}{\lambda _{\max }}({W_l}) \le {M}\} \subseteq \mathbb{R}^{S},
	\end{align*}
	we have
	\begin{equation*}
		\RadeN(\nn(\Width,\Depth); \data)\le \frac{{2r\sqrt \Depth {(aM)^{\Depth}}}}{{\sqrt n }}{\E}\mathop {\max }\limits_{i \in [n]}{\left\| \data_i \right\|_2}.
	\end{equation*}
\end{lemma}

The proof of Lemma~\ref{lem:RadComplexityNN} is given in Appendix.
The term ${\E}\mathop {\max }_{i \in [n]}{\left\| \data_i \right\|_2}=O(\sqrt{\dimInput})$ due to $\data_i\in[0,1]^p$. But we suppose $\dimInput$ is a fixed number in assumption A.1, ${\E}\mathop {\max }_{i \in [n]}{\left\| \data_i \right\|_2}$ could be omitted when $\samplesize$ is large enough.  For a fixed $\Depth$, we can get the order of Rademacher complexity of DNNs is $O(n^{-1/2})$.\\

The following corollary is a straightforward result of Proposition~\ref{prop:genApproxErr} and Lemma~\ref{lem:RadComplexityNN}. By taking $\inf_{\generator\in\genSpace}$, it gives high probability bounds for the third term in the error decomposition lemma.
\begin{corollary}
	\label{coro:genApproxErr}
	Suppose that the conditions in Proposition~\ref{prop:genApproxErr} and Lemma~\ref{lem:RadComplexityNN} hold.
	We have
	\begin{align*}
		&~~~~\inf_{\generator\in\genSpace}\sup_{\discriminator\in\disSpace}\bigl\{\MOM_\blockNum[\discriminator]-\E_{\generator_{\#}\distriN}[\discriminator]\bigr\}\\
		&\le\frac{4{\E}\mathop {\max }\limits_{i \in I}{\left\| \data_i \right\|_2}r\sqrt L_\disSpace {(aM)^{L_\disSpace}}}{{\sqrt n }}+{\frac{32\sqrt K{ }{\E}\mathop {\max }\limits_{i \in B_1}{\left\| \data_i \right\|_2}r\sqrt L_2 {(aM)^{L_\disSpace}}}{[1-(2\eta(\varepsilon))^{-1}]\sqrt n}}\vee \sqrt{ \frac{[(16\varianceDis)\wedge(32B^2/{\rm{e}})] K^2}{[1-(2\eta(\varepsilon))^{-1}]n}}+\lipSupB\sqrt{\frac{\blockNum}{(1-\varepsilon)n}}
	\end{align*}
	with probability at least $1-{\rm{e}}^{-\blockNum/32}-{\rm{e}}^{-\frac{K}{8\eta(\varepsilon)}\left(1-\frac{1}{2\eta(\varepsilon)}\right)^2}$.
\end{corollary}

\section{Numerical experiments}\label{sec:simApp}
In this section, we apply two real datasets on MoM-GAN
to demonstrate its effectiveness of MoM-GAN. We first give a gradient-based algorithm and then show the numerical results on the MNIST and FashionMNIST datasets.

\subsection{Implementation}
In the implementation,  we consider the minimax problem (\ref{def:momgan}), which is defined as:
\begin{equation}\label{def:optimization}
	\min_{\theta_{\generator}}\max_{\theta_{\discriminator}}\Big\{\MOM_{\blockNum}[\discriminator_{\theta_{\discriminator}}]-\frac{1}{w}\sum_{j=1}^{w}d_{\theta_{\discriminator}}(\generator_{\theta_{\generator}}(\noise_j))\Big\}.
\end{equation}
The details for solving the above optimization problem are shown in Algorithm 1. To tackle the unstable training process, we first train the discriminator $h_d$ times to enhance its ability to distinguish real and generated examples. Then, we train the generator one time once the discriminator is trained $h_d$ times. More precisely, for each batch, we sample $w$ real examples (with potential noise) and then compute the gradient $G_{\theta_d}$ of the loss function (\ref{def:optimization}) with respect to parameters $\theta_d$ of the discriminator based on the sampled examples. Since the median operator makes problem \eqref{def:optimization} non-differentiable. Thus, we adopt the MoM-gradient descent \cite{Lecue2020Robust} which computes the gradient only based on the samples in the median block. Then, the gradient $G_{\theta_d}$ is obtained by 
$$
G_{\theta_d} \leftarrow \frac{K}{n}\sum_{i\in B_{med}}\triangledown_{\theta_d}d(\data_i)-\frac{1}{w}\sum_{j=1}^{w}\triangledown_{\theta_d}d(g(\noise_j)),
$$
where 
$B_{med}$ denotes the median block. 

To update $\theta_d$, we utilize the RMSProp algorithm \citep{tieleman2017divide}, which is quite suitable for a vibrant gradient descent process like the training process here. In the end, we implement the \textit{clip} operation on updated parameters $\theta_d$ so as to prevent exploding gradients. The training process of the generator is not novel. Similar to the training of WGAN, we first sample $w$ random noise $\{ \noise_1,...,\noise_w\}$ to generate $w$ examples. Then, we compute the gradient of loss function (\ref{def:optimization}) with respect to $\theta_g$ and then update $\theta_g$ using RMSProp algorithm.

\begin{algorithm}[H]
	\caption{MoM-GAN}
	\label{alg:the_alg}
	\KwIn{ $\alpha$: the learning rate; $K$: the number of blocks for $d$; $c$: the clipping parameter; $h_d$: the number of steps to train the discriminator;  $w$: the batch size. $T$: the training epochs. }
	\hspace*{-0.15in}\quad  \textbf{Initialize} $\quad \theta_d, \theta_g \leftarrow$ Gaussian. \\
	\For{$t = 1,...,T$}{
		\For {$t_d = 1,...,h_{d}$}{
			\mbox{ Sample $w$ examples $\mathbf{X}=\{\data_{1},...,\data_{w}\}$ from the generating distribution $\mu$;}\\
			\mbox{ Sample $w$ noise samples $\{\noise_{1},...,\noise_{w}\}$ from the noise distribution $\nu$;} \\
			\mbox{ Sample $K$ disjoint blocks $B_1,...,B_K$, and find the median block $B_{med}$ based on $\MOM_{\blockNum}[\discriminator_{\theta_{\discriminator}}]$ of $\mathbf{X}$;}\\
			\mbox{ Update the parameters of the discriminator as follows:}\\
			\mbox{ \hspace{0.15in} $G_{\theta_d}\gets\frac{K}{n}\sum_{i\in B_{med}}\triangledown_{\theta_d}d(\data_i)-\frac{1}{w}\sum_{j=1}^{w}\triangledown_{\theta_d}d(g(\noise_j))$;}\\
			\mbox{ \hspace{0.2in}$\theta_d\gets \theta_d-\alpha\times {\rm{RMSProp}}(\theta_d, G_{\theta_d})$;}\\
			\mbox{ \hspace{0.2in}$\theta_d\gets {\rm{clip}}(\theta_d,-c,c)$.}
		}
		
		\mbox{ Sample $w$ noise samples $\{\noise_{1},...,\noise_{w}\}$ from the noise distribution $\nu$;}\\
		\mbox{ Update the parameters of the generator as follows:} \\
		\mbox{ \hspace{0.2in} $G_{\theta_g}\gets-\triangledown_{\theta_g}\frac{1}{w}\sum_{j=1}^{w}d(g(\noise_j))$}\\
		\mbox{ \hspace{0.2in} $\theta_g\gets \theta_g-\alpha\times{\rm{RMSProp}}(\theta_g,G_{\theta_g})$}
	}
\end{algorithm}

\subsection{Real Data Applications} \label{sec:RealData}
We illustrate the robustness of MoM-GAN by learning the distribution of two image datasets, MNIST in \cite{lecun-mnisthandwrittendigit-2010}, and FashionMNIST in \cite{xiao2017/online}, contaminated by random noisy images and real noisy images. The MNIST dataset is a ten-class handwritten digits database including 70000 $28\times28$ grayscale images. Here 60000 samples are used for training and other 10000 samples for testing. FashionMNIST dataset is also composed of ten classes of 70000 $28\times28$ grayscale Zalando's article images, which is divided into 60000 training data and 10000 testing data. We adopt the widely used GAN structure, DCGAN \cite{DBLP:journals/corr/RadfordMC15}, as the networks of discriminator and generator. To evaluate the robustness of the distribution estimation, we contaminate the MNIST dataset by injecting noisy images into the training set, in which pixels of each noisy image are sampled from the Gaussian distribution. More precisely, we randomly contaminate each batch with a probability 50\%, by incorporating $\pi \times w$ Gaussian noisy images into each polluted batch, where $w=64$ is the batch size and $\pi$ is the proportion of noisy images. To illustrate the impact of the number of noisy images on density estimation performance, we consider two different settings of $\pi$ being $2\%$ and $4\%$, respectively. The number of blocks, i.e., $K$ in Algorithm 1, is equal to 4, and the epoch $T$ in Algorithm 1 is set to be 200. The details about the selection of $K$ will be discussed in subsection \ref{sec:block}. For the FashionMNIST dataset, we adopt the same way to contaminate each batch with probability $50\%$ by the real noisy images. In other words, we inject the $4$th class image in MNIST (the handwritten digit 3) into FashinMNIST. The proportion of noisy images $\pi$ is set $2\%$ and $4\%$. The number of blocks $K$ is 8. 

\begin{figure}[!t]
	\centering
	% include first image
	\subfloat[MoM-GAN]{
		\includegraphics[width=0.35\linewidth]{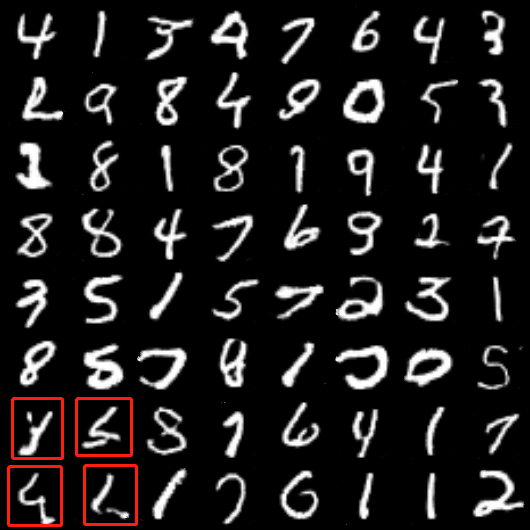}}\quad
	\subfloat[WGAN]{
		\includegraphics[width=0.35\linewidth]{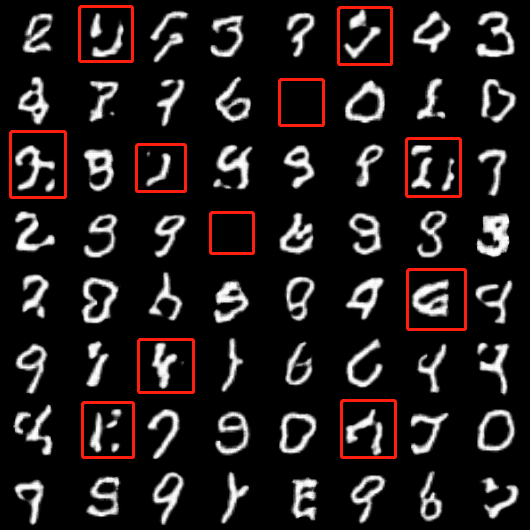}}\\
	\caption{{Generated samples from trained MoM-GAN and WGAN for adding 2\% noise proportion Gaussian noisy images on each batch for MNIST dataset (abnormal images are highlighted with red boxes).}}
	\label{fig:mnist0.02}
\end{figure}

\begin{figure}[!t]
	\centering
	% include first image
	\subfloat[MoM-GAN]{
		\includegraphics[width=0.35\linewidth]{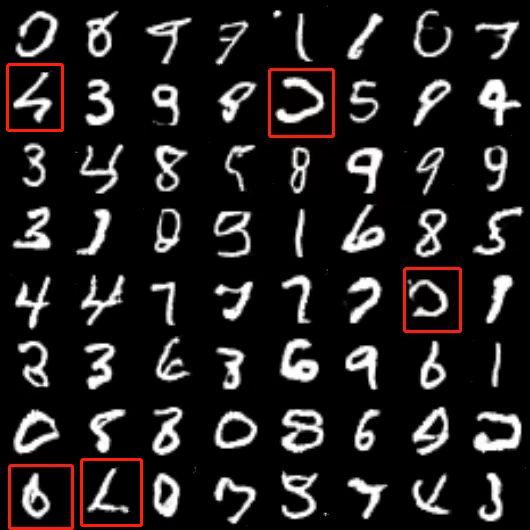}}\quad
	\subfloat[WGAN]{
		\includegraphics[width=0.35\linewidth]{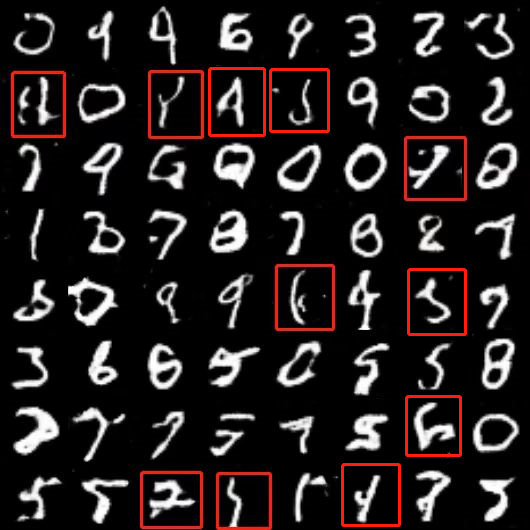}}\\
	\caption{{Generated samples from trained MoM-GAN and WGAN for adding 4\% noise proportion Gaussian noisy images on each batch for MNIST dataset (abnormal images are highlighted with red boxes).}}
	\label{fig:mnist0.04}
\end{figure}

\begin{figure}[!t]
	\centering
	% include second image
	\subfloat[MoM-GAN]{
		\includegraphics[width=0.35\linewidth]{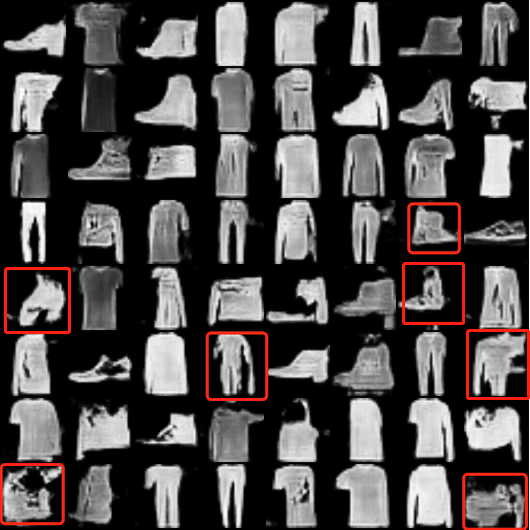}}\quad
	\subfloat[WGAN]{
		% include second image
		\includegraphics[width=0.35\linewidth]{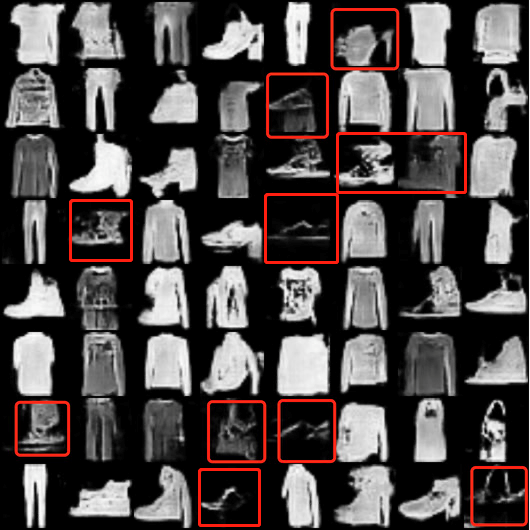}}\\
	\caption{{Generated samples from trained MoM-GAN and WGAN for adding 2\% real noisy images on each batch for FashionMNIST dataset (abnormal images are highlighted with red boxes).}}
	\label{fig:fashionmnist0.02}
\end{figure}

\begin{figure}[!t]
	\centering
	% include second image
	\subfloat[MoM-GAN]{
		\includegraphics[width=0.35\linewidth]{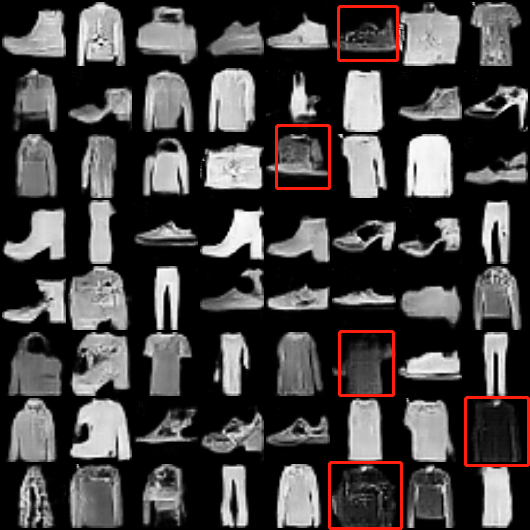}}\quad
	\subfloat[WGAN]{
		\includegraphics[width=0.35\linewidth]{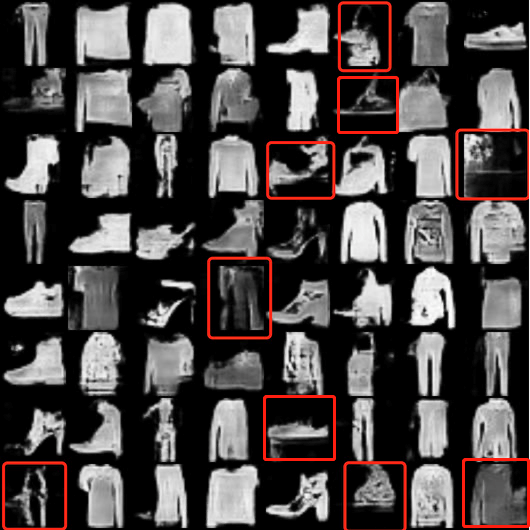}}\\
	\caption{{Generated samples from trained MoM-GAN and WGAN for adding 4\% real noisy images on each batch for FashionMNIST dataset (abnormal images are highlighted with red boxes).}}
	\label{fig:fashionmnist0.04}
\end{figure}

Figures \ref{fig:mnist0.02}-\ref{fig:mnist0.04} present the images generated by MoM-GAN and WGAN trained on the polluted MNIST dataset with Gaussian noisy images. When the proportion of noisy images incorporated into the training dataset increases from $2\%$ to $4\%$ (Figures \ref{fig:mnist0.02}a-\ref{fig:mnist0.04}a), MoM-GAN performs quite stably by merely producing few abnormal MNIST images. In comparison, WGAN produces more abnormal images, as shown in Figure \ref{fig:mnist0.02}b. Such an unnatural phenomenon reveals that the traditional WGAN not only estimates the distribution of normal images but also learns the distribution of noisy images. Therefore, we can conclude that the conventional WGAN is not robust for the contaminated training dataset. In comparison, MoM-GAN can be almost disturbed by a few noisy images and robustly estimate the distribution of real images. The same conclusion is also tenable for MoM-GAN trained on polluted FashionMNIST with real noisy images, as shown in Figures \ref{fig:fashionmnist0.02}-\ref{fig:fashionmnist0.04}.

For quantitatively demonstrating the robustness of MoM-GAN, we adopt the Fr${\rm{\acute{e}}}$chet Inception Distance (FID, \citep{dowson1982frechet}) score to evaluate the difference between the distribution of real images and that of images generated by MoM-GAN and WGAN. The FID score is defined as follows.
$$
{\rm{FID}}:= \|\mathbf{m}_1-\mathbf{m}_2\|_2^2+{\rm{tr}}(\mathbf{\Sigma}_1+\mathbf{\Sigma}_2-2(\mathbf{\Sigma}_1\mathbf{\Sigma}_2)^{\frac{1}{2}}),
$$
where $\mathbf{m}_1$ and $\mathbf{m}_2$ are the deep feature-wise means of real images and generated images, $\mathbf{\Sigma}_1$ and $\mathbf{\Sigma}_2$ are the feature-wise covariance matrices of real images and generated images. The smaller the FID score is, the distribution of generated images is closer to the real one. In practice, we compute the FID scores according to 10000 real images in the test dataset and 10000 fake images generated by MoM-GAN and WGAN. 

Tables \ref{tab:FID-MNIST-G} and \ref{tab:FID-MNIST-P} summarize the FID scores of generated images of MoM-GAN and WGAN under different settings of noise proportions $\{2\%,  4\%\}$ on each batch for MNIST dataset. Table \ref{tab:FID-fashion} presents the results of FashionMNIST dataset with $\{2\%, 4\%\}$ noise proportion on each batch. As we can see, the reported FID scores are coincided with the visual results presented in Figures \ref{fig:mnist0.02}-\ref{fig:fashionmnist0.04}. Taking MNIST (Table \ref{tab:FID-MNIST-G}) as an example, as the noise proportion rises, the FID score of WGAN gradually increases, which implies that there exists a large distance between the distribution of images generated by WGAN and real images. Visually, as shown in Figures \ref{fig:mnist0.02}b-\ref{fig:mnist0.04}b, several images generated by WGAN are quite distinct from the real MNIST images. In comparison, the FID scores of MoM-GAN are quite stable, which reflects that the MoM-GAN is robust to the polluted examples in terms of distribution estimation. We also train the MoM-GAN and WGAN on polluted MNIST dataset with Pareto distributed noisy images (scale parameter is 1 and shape parameter is 2), see Table \ref{tab:FID-MNIST-P}. It is well known that the Pareto distribution is heavy-tailed, which is 
skew distribution different from the Gaussian distribution. The results of Table \ref{tab:FID-MNIST-P} indicate that the FID scores of MoM-GAN are smaller than that of WGAN, which implies that the robustness of MoM-GAN also exits for the polluted data with uncommon random noise. A similar conclusion can be concluded from Table \ref{tab:FID-fashion}  in the case of estimating the distribution of the FashionMNIST dataset by MoM-GAN and WGAN.

In Section \ref{sec:theory}, our main results are derived from the assumption that less than half of the blocks are contaminated by noise. Here we empirically verify that under this assumption, the robustness of MoM-GAN becomes weak following the increase of the number of polluted blocks. The comparing results are recorded in Table \ref{tab:FID-MNIST-U-G}-\ref{tab:FID-fashion-U-R}. The noise type,  30\% blocks and 50\% blocks represent 30\% or 50\% blocks are contaminated. As can be seen, the FID scores of MoM-GAN based on less polluted blocks (30\% blocks) are much smaller than that of MoM-GAN based on more polluted blocks (50\% blocks). We note that the MoM-GAN in these two cases still outperforms than the traditional WGAN.

\subsection{Selection of Blocks}\label{sec:block}
\begin{figure}[!t]
		\centering
		% include first image
		\subfloat{
			\includegraphics[width=0.48\linewidth]{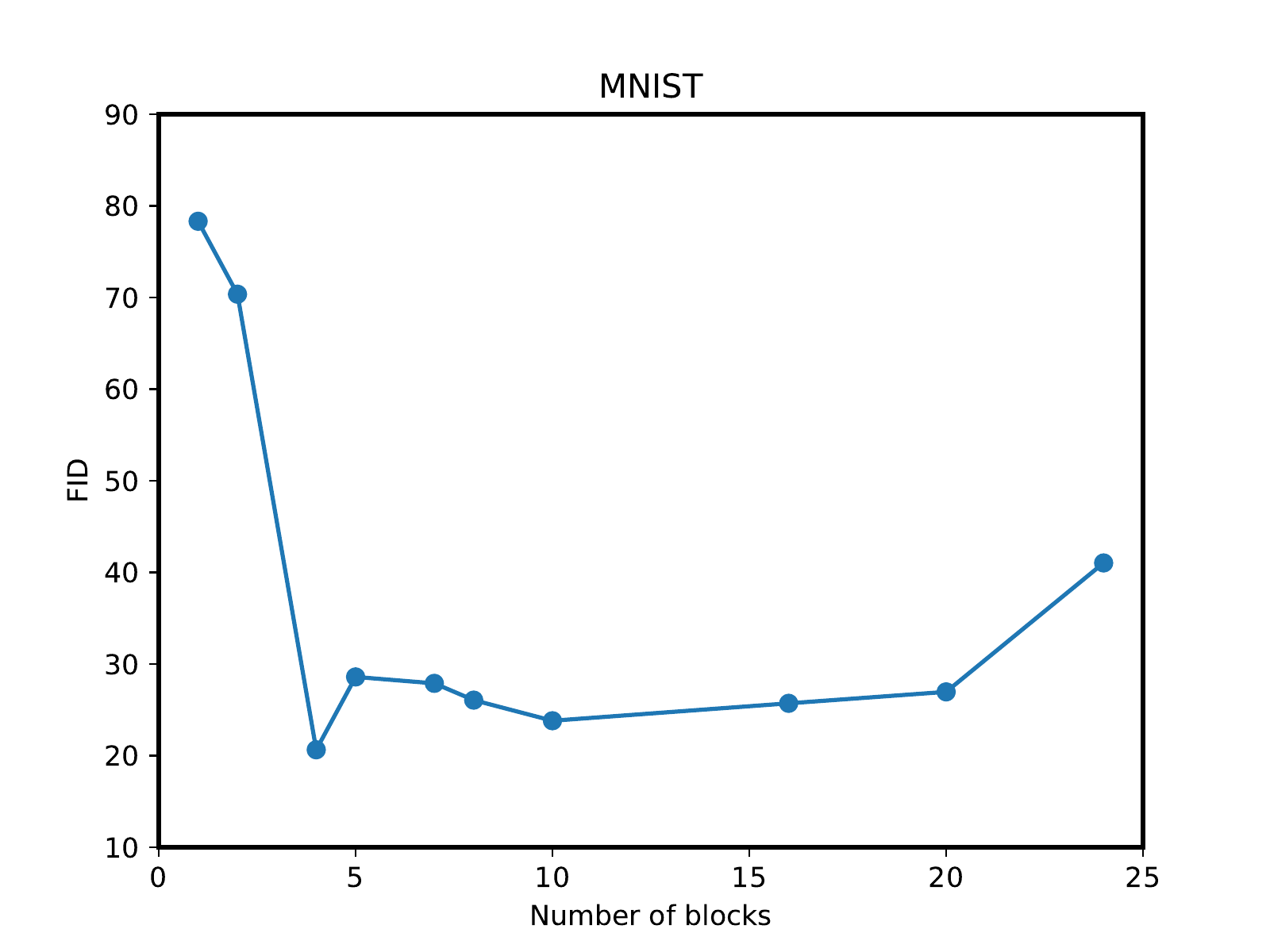}}\quad
		\subfloat{
			\includegraphics[width=0.48\linewidth]{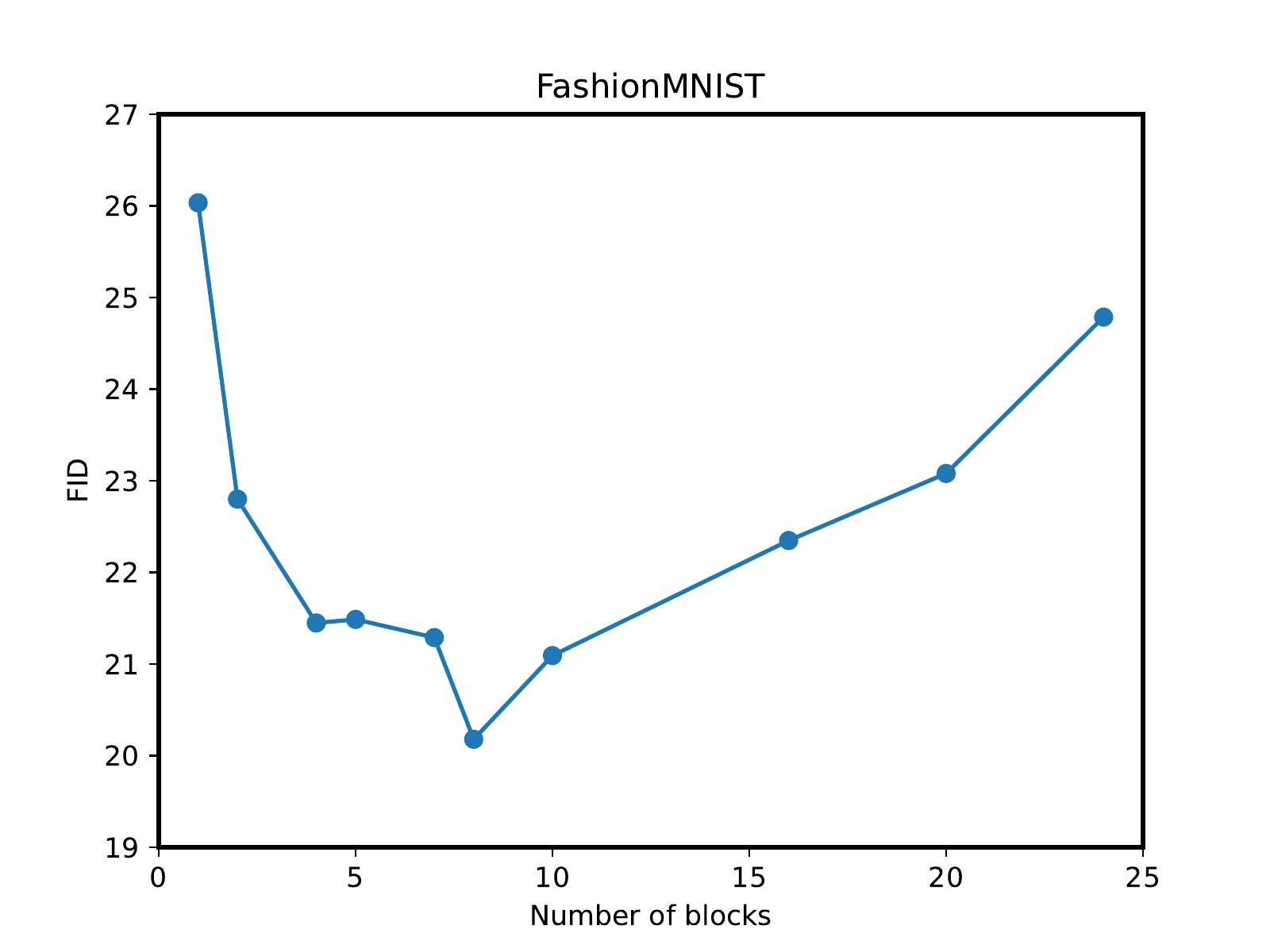}}\\
		\caption{{FID changes on polluted datasets with $4\%$ noisy images under different blocks. Left: MNIST contaminated by Gaussian noisy images. Right: FashionMNIST contaminated by real noisy images.}}
		\label{fig:blocks}
	\end{figure}

The number of blocks $K$ is a crucial parameter that directly impacts the robustness of MoM-GAN. Figure \ref{fig:blocks} plots the impact of $K$ on the FID scores of image generated by MoM-GAN. Intuitively, the two curves rapidly decrease to the minimal point and slowly rise. In detail, when $K=1$, MoM-GAN loss reduces to the mean of Wasserstein distance, and naturally the MoM-GAN is equivalent to the WGAN. Thus, in this case, images (e.g., MNIST polluted by Gaussian noisy images) generated by MoM-GAN ($K=1$) and WGAN achieve the similar FID scores, i.e., $79.72$ for MoM-GAN (see Figuer \ref{fig:blocks}) and $78.32$ for WGAN (see Table \ref{tab:FID-MNIST-G}). At the other extreme that $K$ tends to a large value (e.g., 24), the performance of MoM-GAN is even worse than that of models with a small block number $K$. The reason is that a large value of $K$ would decrease the number of observations used for estimating the distribution of the dataset. Taking the distribution estimation of the MNIST as an example, when $K=24$ and batch size $w=64$, each block contains two images, which implies 62 observations will not be considered to estimate the distribution of MNIST. From the perspective of the whole dataset, merely $60000 \times 3/64$ examples are involved in training the MoM-GAN), naturally resulting in unsatisfactory performance. Therefore, in practice, we suggest searching $K$ in the interval $[2,\lfloor w/6\rfloor]$ so as to avoid removing too many observations or reducing the MoM-GAN to WGAN. Following this guideline, we adopt the binary search in the interval $[2,10]$ and finally determine $K=4$ for MNIST and $K=8$ for FashionMNIST, as introduced in subsection \ref{sec:RealData}.

\begin{table}[!t]
	\centering
	\caption{FID on polluted MNIST dataset with Gaussian distributed noisy images.}
	\setlength{\tabcolsep}{7mm}{
		%	\resizebox{\textwidth}{10mm}{
			\begin{tabular}{cccccc}
				\toprule
				Noise proportion ($\pi$) &       & MoM-GAN &       &       & WGAN \\
				\midrule
				2\%  &       &\textbf{18.55} &       &       & 76.51\\
				4\%  &       &  \textbf{20.64} &       &       & 79.72\\
				\bottomrule
		\end{tabular}}%
		\label{tab:FID-MNIST-G}%
	\end{table}%
	
	\begin{table}[!tp]
		\centering
		\caption{FID on polluted MNIST dataset with Pareto distributed noisy images.}
		\setlength{\tabcolsep}{7mm}{
			\begin{tabular}{cccccc}
				\toprule
				Noise proportion ($\pi$)&       & MoM-GAN &       &       & WGAN \\
				\midrule
				2\%  &       & \textbf{20.89} &       &       &33.71\\
				4\%  &       &  \textbf{20.17} &       &       &45.92 \\
				\bottomrule
		\end{tabular}}%
		\label{tab:FID-MNIST-P}%
	\end{table}%
	
	\begin{table}[!tp]
		\centering
		\caption{FID on polluted FashionMNIST dataset with real noisy images.}
		\setlength{\tabcolsep}{7mm}{
			\begin{tabular}{cccccc}
				\toprule
				Noise proportion ($\pi$)&       & MoM-GAN &       &       & WGAN \\
				\midrule
				2\%  &       & \textbf{20.17} &       &       & 23.75\\
				4\%  &       &  \textbf{21.48} &       &       &25.61 \\
				\bottomrule
		\end{tabular}}%
		\label{tab:FID-fashion}%
	\end{table}%

	\begin{table}[!tp]
		\centering
		\caption{FID on polluted MNIST dataset with $10\%$ Gaussian distributed noisy images.}
		\setlength{\tabcolsep}{7mm}{
			\begin{tabular}{cccccccc}
				\toprule
				Method &       & Noise type &       & FID  \\
				\midrule
				MoM-GAN &       & 30\% blocks  &       &\textbf{21.61}  \\
				MoM-GAN &       &50\% blocks  &       &  25.11  \\ WGAN &       &------  &       &  93.43 \\
				\bottomrule
		\end{tabular}}%
		\label{tab:FID-MNIST-U-G}%
	\end{table}%
	
	\begin{table}[!tp]
		\centering
		\caption{FID on polluted MNIST dataset with $10\%$ Pareto distributed noisy images.}
		\setlength{\tabcolsep}{7mm}{
			\begin{tabular}{cccccccc}
				\toprule
				Method &       & Noise type &       & FID  \\
				\midrule
				MoM-GAN &       & 30\% blocks  &       & \textbf{24.93} \\
				MoM-GAN &       &50\% blocks  &       & 30.27  \\ 
				WGAN &       &------  &       &  87.99 \\
				\bottomrule
		\end{tabular}}%
		\label{tab:FID-MNIST-U-P}%
	\end{table}%
	
	\begin{table}[!tp]
		\centering
		\caption{FID on polluted FashionMNIST dataset with $13\%$ real noisy images.}
		\setlength{\tabcolsep}{7mm}{
			\begin{tabular}{cccccccc}
				\toprule
				Method &       & Noise type &       & FID  \\
				\midrule
				MoM-GAN &       &  30\% blocks  &       & \textbf{23.92} \\
				MoM-GAN &       &50\% blocks  &       & 24.31  \\ 
				WGAN &       &------  &       &  118.09\\
				\bottomrule
		\end{tabular}}%
		\label{tab:FID-fashion-U-R}%
	\end{table}%

	\section{Conclusion}\label{sec:conclusion}
	This paper studies the MoM-GAN method to estimate an unknown probability measure $\distriT$ when data are contaminated adversarially. Suppose that the generator and discriminator are both constructed by deep ReLU networks, and we give the non-asymptotic IPM error bound for our estimator in Theorem~\ref{thm:main}, which depends on the sample size, the width and depth of the neural network, and the input dimension. 
	To implement our method, we develop a gradient-based descent algorithm (see Algorithm~\ref{alg:the_alg}). 
	We study two popular image datasets, MNIST and FashionMNIST, to examine our method and compare it with the WGAN method. The results show that the trained images by our approach have higher resolution and lower FID than that of WGAN. 
	
	The number of blocks $\blockNum$ is a crucial parameter for the MoM-based method. If $\blockNum$ is too large, the computational cost will be high, and the upper bound of the evaluation error may not converge (see Theorem~\ref{thm:main}). On the other hand, if $\blockNum$ is too small, the robustness will be reduced, and the probability bound will be poor (see Theorem~\ref{thm:main}). The cross-validation scheme may help to calibrate the parameter $\blockNum$, but its heavy computational consumption is also a problem. So, developing an efficient and adaptive calibration scheme for the number of blocks should be future work.

	\section{Proofs}\label{sec:proofs}

	\noindent\textbf{Proof of Theorem~\ref{thm:main} }
	First, we bound $\mathcal{E}\left(\mathcal{H}, \disSpace,[0,1]^{\dimInput}\right)$.
	We construct a neural network $\phi$ to approximate any given function in $\mathcal{H}^{b}\left([0,1]^{\dimInput}\right)$.
	
	For the discriminator $\discriminator\in\mathcal{N} \mathcal{N}\left(\disWidth,\disDepth\right) \cap \operatorname{Lip}\left(\mathbb{R}^{p}, J, 1\right)$, Lemma 2.7 in \citet{huang2021error} guarantees that, for a given $f \in \mathcal{H}^{b}\left([0,1]^{p}\right)$, there exists a neural network $\phi$ with
	\begin{center}
		width $49(s+1)^{2} 3^{p} p^{s+1} N_{\disSpace}\left\lceil\log_{2} N_{\disSpace}\right\rceil$ and depth $15(s+1)^{2} L_{\disSpace}\left\lceil\log _{2} L_{\disSpace}\right\rceil+2 p$
	\end{center}
	such that $\phi \in \operatorname{Lip}\left(\mathbb{R}^{p}, J, 1\right)$ with Lipschitz constant
	$J \le (s+1) p^{s+1 / 2} L_{\disSpace}(\disWidth\disDepth)^{{\color{black}{((4 {b}-4)\vee 0)}} / p}(1260 \disWidth^{2}\disDepth^{2} 2^{L_{\disSpace}^{2}}+19 s 7^{s})$
	and
	\begin{center}
		$\|\phi-f\|_{L^{\infty}([0,1]^{p})} \leq 6(s+1)^{2} p^{(s+{b} / 2) \vee 1}\left\lfloor(\disWidth\disDepth)^{2 / p}\right\rfloor^{-{b}} .$
	\end{center}
	For the given sample sive $n$, if we choose $\disWidth$ and $\disDepth$ such that $C_p n^{1 / 2} \le \disWidth\disDepth $ for some constant $C_p>0$, by the setting of discriminator class above, one has
	\begin{align*}
		\mathcal{E}\left(\mathcal{H}, \disSpace,[0,1]^{p}\right)& =\sup_{\phi\in\evaSpace}\inf_{f\in\disSpace}\|\phi-f\|_{L^{\infty}([0,1]^{p})}\\
		&\le 6(s+1)^{2} p^{(s+{b} / 2) \vee 1}\left\lfloor(\disWidth\disDepth)^{2 / p}\right\rfloor^{-{b}} \\
		&\le 6(s+1)^{2} p^{(s+{b} / 2) \vee 1}\left[C_p^{2 / p}n^{1/ p}-1\right]^{-{b}} .
	\end{align*}
	
	Note that $\distance_{\disSpace \circ \genSpace}(\widehat{\nu}_{m}, \nu)=\sup_{\discriminator\composition\generator\in\disSpace\composition\genSpace}\bigl|\frac{1}{\samplesizeNoise}\sum_{i=1}^{\samplesizeNoise}\bigl((\discriminator\composition\generator)(\noise_i)-\E_{\noise_i\sim\distriN}[(\discriminator\composition\generator)(\noise_i)]\bigr)\bigr|$. 
	We can obtain $\E \distance_{\disSpace \circ \genSpace}(\widehat{\nu}_{m}, \nu)\le 2\Rade_{\samplesizeNoise}(\disSpace\composition\genSpace;\noise)$ by the symmetrization theorem (see e.g. Lemma 2.3.1 in \cite{vaart1996weak}). Under assumption $\sup_{\discriminator\in\disSpace}\normsup{\discriminator}\le\lipSupB$, Massart's concentration inequality for suprema of the bounded empirical processes [see Theorem~9 in~\cite{massart2000constants}] shows that
	\begin{align*}
		\PP\biggl\{\distance_{\disSpace \circ \genSpace}(\widehat{\nu}_{m}, \nu)\ge2\Rade_{\samplesizeNoise}(\disSpace\composition\genSpace)+2\lipSupB\sqrt{\frac{8t}{\samplesizeNoise}}\biggr\}\le\PP\biggl\{\distance_{\disSpace \circ \genSpace}(\widehat{\nu}_{m}, \nu)\ge\E \distance_{\disSpace \circ \genSpace}(\widehat{\nu}_{m}, \nu)+2\lipSupB\sqrt{\frac{8t}{\samplesizeNoise}}\biggr\}\le {\rm{e}}^{-t}.
	\end{align*}
	
	Next, we need to calculate $\Rade_{\samplesizeNoise}(\disSpace\composition\genSpace)$. Notice that the composition of two neural networks $\discriminator\in\disSpace=\nn(\disWidth,\disDepth)$ and $\generator\in\genSpace=\nn(\genWidth,\genDepth)$ is still a neural network with width $\max\{\disWidth,\genWidth\}$ and depth $\disDepth+\genDepth+1$. Lemma~\ref{lem:RadComplexityNN} implies
	\begin{align*}
		\E \distance_{\disSpace \circ \genSpace}(\widehat{\nu}_{m}, \nu)\le 2\Rade_{\samplesizeNoise}(\disSpace\composition\genSpace;\noise)\le\frac{{4r\sqrt{\disDepth+\genDepth+1} {(aM)^{\disDepth+\genDepth+1}}}}{{\sqrt\samplesizeNoise}}{\E}\mathop {\max }\limits_{i\in [\samplesizeNoise]}{\left\| \noise_i\right\|_2}.
	\end{align*}
	
	Thus, with probability at least $1-{\rm{e}}^{-t}$,
	$$\distance_{\disSpace \circ \genSpace}(\widehat{\nu}_{m}, \nu)\le \frac{{4r\sqrt{\disDepth+\genDepth+1} {(aM)^{\disDepth+\genDepth+1}}}}{{\sqrt\samplesizeNoise}}{\E}\mathop {\max }\limits_{i\in [\samplesizeNoise]}{\left\| \noise_i\right\|_2}+2\lipSupB\sqrt{\frac{8t}{\samplesizeNoise}}.$$
	
	%\red{One comment: Replace $\Width_1,\Depth_1$ by $\genWidth,\genDepth$. Also for $\disWidth\disDepth$. }
	
	Second, Let $B=1$ and $a=1$. By Lemma~\ref{lem:errDecomposition}, \eqref{eq:Lerasle2019} in Appendix and  Proposition \ref{prop:genApproxErr}, we have 
	\begin{align*}
		&~~~~\distance_{\evaSpace}(\distriT,(\genEst_{\momgan})_{\#}\distriN)\le 2\approxErr(\evaSpace,\disSpace,\domain)+\sup_{\discriminator\in\disSpace}\bigl\{|\E_{\distriT}[\discriminator]-\MOM_{\blockNum}[\discriminator]|\bigr\}+\inf_{\generator\in\genSpace}\sup_{\discriminator\in\disSpace}\{\MOM_{\blockNum}[\discriminator]-\E_{\generator_{\#}\distriN}[\discriminator]\}+2\distance_{\disSpace \circ \genSpace}(\widehat{\nu}_{m}, \nu)\\
		&\le \frac{12(s+1)^{2} p^{(s+{b} / 2) \vee 1}}{[C_p^{2 /p}n^{1/ p}-1]^{b}}
		+\frac{16\Rade_{n/K}(\disSpace; \data) }{1-(2\eta(\varepsilon))^{-1}} \vee \sqrt{ \frac{[(16\varianceDis)\wedge(32/{\rm{e}})] K^2}{[1-(2\eta(\varepsilon))^{-1}]n}}+2\Rade_{(1-\varepsilon) n}(\disSpace; \data)\\
		&+\frac{16\Rade_{n/K}(\disSpace; \data) }{1-(2\eta(\varepsilon))^{-1}}\vee \sqrt{ \frac{[(16\varianceDis)\wedge(32/{\rm{e}})] K^2}{[1-(2\eta(\varepsilon))^{-1}]n}}+\sqrt{\frac{\blockNum}{(1-\varepsilon)n}}+\frac{{4r\sqrt{\disDepth+\genDepth+1} {M^{\disDepth+\genDepth+1}}}}{{\sqrt\samplesizeNoise}}{\E}\mathop {\max }\limits_{i\in [\samplesizeNoise]}{\left\| \noise_i\right\|_2}+2\sqrt{\frac{8t}{\samplesizeNoise}}\\
		&\le\frac{12(s+1)^{2} p^{(s+{b} / 2) \vee 1}}{[C_p^{2 /p}n^{1/ p}-1]^{b}}+2\Rade_{(1-\varepsilon)\samplesize}(\disSpace; \data)+\frac{32\Rade_{n/K}(\disSpace; \data) }{1-(2\eta(\varepsilon))^{-1}}\vee 8\sqrt{ \frac{[\varianceDis\wedge(2/{\rm{e}})] K^2}{[1-(2\eta(\varepsilon))^{-1}]n}}\\
		& +\frac{{4r\sqrt{\disDepth+\genDepth+1} {(aM)^{\disDepth+\genDepth+1}}}}{{\sqrt\samplesizeNoise}}{\E}\mathop {\max }\limits_{i\in [\samplesizeNoise]}{\left\| \noise_i\right\|_2}+2\sqrt{\frac{8t}{\samplesizeNoise}}\\
		& \le \frac{12(s+1)^{2} p^{(s+{b} / 2) \vee 1}}{[C_p^{2 / p}n^{1/ p}-1]^{b}}+\frac{4{\E}\mathop {\max }\limits_{i \in I}{\left\| \data_i \right\|_2}r\sqrt \disDepth {M^{\disDepth}}}{{\sqrt n }}+{\frac{64\sqrt K{ }{\E}\mathop {\max }\limits_{i \in B_1}{\left\| \data_i \right\|_2}r\sqrt \disDepth {M^{\disDepth}}}{[1-(2\eta(\varepsilon))^{-1}]\sqrt n}}\vee \sqrt{ \frac{[(16\varianceDis)\wedge(32/{\rm{e}})] K^2}{[1-(2\eta(\varepsilon))^{-1}]n}}\\
		& +\frac{{4r\sqrt{\disDepth+\genDepth+1} {M^{\disDepth+\genDepth+1}}}}{{\sqrt\samplesizeNoise}}{\E}\mathop {\max }\limits_{i\in [\samplesizeNoise]}{\left\| \noise_i\right\|_2}+2\sqrt{\frac{8t}{\samplesizeNoise}}+\sqrt{\frac{\blockNum}{(1-\varepsilon)n}}\\
		& =O(n^{-{b} / p})+ O\left(\frac{1}{\sqrt{n}}{[rM^{\disDepth}}\sqrt \disDepth+\sqrt{\frac{\blockNum}{(1-\varepsilon)}}]\right)+O\left(\sqrt{\frac{K\disDepth}{n}}\frac{M^{\disDepth}}{[1-(2\eta(\varepsilon))^{-1}]}\right)\vee O\left(\sqrt{\frac{K^2}{[1-(2\eta(\varepsilon))^{-1}]n}}\right)\\
		&+O\left(r\sqrt{\frac{\disDepth+\genDepth}{m}}{M^{{\disDepth+\genDepth}}}+\sqrt{\frac{t}{m}}\right),
	\end{align*}
	with probability at least $1-{\rm{e}}^{-\blockNum/32}-2{\rm{e}}^{-\frac{K}{8\eta(\varepsilon)}\left(1-\frac{1}{2\eta(\varepsilon)}\right)^2}-{\rm{e}}^{-t}$.
	%where we use $x\vee y \le x+y$ in the inequality for $x,y \ge 0$.
	
	\bigskip
	
	%\newpage
	%\begin{appendices}
	\noindent\textbf{Proof of Lemma~\ref{lem:errDecomposition} }
	For any $\epsilon>0$, we can find a $\fcnEva_{\epsilon}\in\evaSpace$ such that
	\begin{align*}
		\distance_{\evaSpace}(\distriT,\generator_{\#}\distriN)
		&=\sup_{\fcnEva\in\evaSpace}\bigl\{\E_{\distriT}[\fcnEva]-\E_{\generator_{\#}\distriN}[\fcnEva]\bigr\}\le\E_{\distriT}[\fcnEva_{\epsilon}]-\E_{\generator_{\#}\distriN}[\fcnEva_{\epsilon}]+\epsilon.
	\end{align*}
	Consequently, we have
	\begin{align*}
		\distance_{\evaSpace}(\distriT,\generator_{\#}\distriN)	&\le\E_{\distriT}[\fcnEva_{\epsilon}-\discriminator_\epsilon]+\E_{\distriT}[\discriminator_\epsilon]-\MOM_{\blockNum}[\discriminator_\epsilon]+\MOM_{\blockNum}[\discriminator_\epsilon]-\E_{\generator_{\#}\distriN}[\discriminator_\epsilon]+\E_{\generator_{\#}\distriN}[\discriminator_\epsilon-\fcnEva_{\epsilon}]+\epsilon\\
		&\le 2\|\fcnEva_{\epsilon}-\discriminator_\epsilon\|_{L^{\infty}(\domain)}+\bigl(\E_{\distriT}[\discriminator_\epsilon]-\MOM_{\blockNum}[\discriminator_\epsilon]\bigr)+\bigl(\MOM_{\blockNum}[\discriminator_\epsilon]-\E_{\generator_{\#}\distriN}[\discriminator_\epsilon]\bigr)+\epsilon\\
		[\text{By \eqref{eq:error} below}~]&\le2\inf_{\discriminator\in\disSpace}\|\fcnEva_{\epsilon}-\discriminator\|_{L^{\infty}(\domain)}+\sup_{\discriminator\in\disSpace}\bigl\{\E_{\distriT}[\discriminator]-\MOM_{\blockNum}[\discriminator]\bigr\}+\bigl(\MOM_{\blockNum}[\discriminator_\epsilon]-\E_{\generator_{\#}\distriN}[\discriminator_\epsilon]\bigr)+3\epsilon.
	\end{align*}
	where for given $\epsilon$ and $\fcnEva_{\epsilon}$, we can further find a $\discriminator_\epsilon$ such that
	\begin{equation}\label{eq:error} \|\fcnEva_{\epsilon}-\discriminator_\epsilon\|_{L^{\infty}(\domain)}\le\inf_{\discriminator\in\disSpace}\|\fcnEva_{\epsilon}-\discriminator\|_{L^{\infty}(\domain)}+\epsilon.
	\end{equation}
	
	Letting $\generator=\genEst_{\momgan}$ and $\approxErr(\evaSpace,\disSpace,\domain):=\sup_{\fcnEva\in\evaSpace}\inf_{\discriminator\in\disSpace}\|\fcnEva-\discriminator\|_{L^\infty(\domain)}$, we have
	\begin{align*}
		\distance_{\evaSpace}(\distriT,(\genEst_{\momgan})_{\#}\distriN)& \le 2\inf_{\discriminator\in\disSpace}\|\fcnEva_{\epsilon}-\discriminator\|_{L^{\infty}(\domain)}+\sup_{\discriminator\in\disSpace}\bigl\{\E_{\distriT}[\discriminator]-\MOM_{\blockNum}[\discriminator]\bigr\}+\bigl(\MOM_{\blockNum}[\discriminator_\epsilon]-\E_{(\genEst_{\momgan})_{\#}\distriN}[\discriminator_\epsilon]\bigr)+3\epsilon\\
		&= 2\inf_{\discriminator\in\disSpace}\|\fcnEva_{\epsilon}-\discriminator\|_{L^{\infty}(\domain)}+\sup_{\discriminator\in\disSpace}\bigl\{\E_{\distriT}[\discriminator]-\MOM_{\blockNum}[\discriminator]\bigr\}+\biggl(\MOM_{\blockNum}[\discriminator_\epsilon]-{\frac{1}{\samplesizeNoise}\sum_{i=1}^{\samplesizeNoise}\discriminator_{\epsilon}(\genEst_{\momgan}(\noise_i))}\biggr)\\
		&+\biggl({\frac{1}{\samplesizeNoise}\sum_{i=1}^{\samplesizeNoise}\discriminator_{\epsilon}(\genEst_{\momgan}(\noise_i))} -\E_{(\genEst_{\momgan})_{\#}\distriN}[\discriminator_\epsilon]\biggr)+3\epsilon.
	\end{align*}
	
	By the definitions of $\genEst_{\momgan}$ in~\eqref{def:momgan}, the previous bound can be further bounded by
	\begin{align*}
		&~~~~\distance_{\evaSpace}(\distriT,(\genEst_{\momgan})_{\#}\distriN) \le 2\approxErr(\evaSpace,\disSpace,\domain)+\sup_{\discriminator\in\disSpace}\bigl\{\E_{\distriT}[\discriminator]-\MOM_{\blockNum}[\discriminator]\bigr\}+\inf_{\generator\in\genSpace}\sup_{\discriminator\in\disSpace}\{\MOM_{\blockNum}[\discriminator]-\frac{1}{\samplesizeNoise}\sum_{i=1}^{\samplesizeNoise}\discriminator(\generator(\noise_i))\}\\
		&+\biggl({\frac{1}{\samplesizeNoise}\sum_{i=1}^{\samplesizeNoise}\discriminator_{\epsilon}(\genEst_{\momgan}(\noise_i))} -\E_{(\genEst_{\momgan})_{\#}\distriN}[\discriminator_\epsilon]\biggr)+3\epsilon\\
		&\le 2\approxErr(\evaSpace,\disSpace,\domain)+\sup_{\discriminator\in\disSpace}\bigl\{\E_{\distriT}[\discriminator]-\MOM_{\blockNum}[\discriminator]\bigr\}+\inf_{\generator\in\genSpace}\sup_{\discriminator\in\disSpace}\{\MOM_{\blockNum}[\discriminator]-\frac{1}{\samplesizeNoise}\sum_{i=1}^{\samplesizeNoise}\discriminator(\generator(\noise_i))\}\\
		&+{\sup_{\discriminator\composition\generator\in\disSpace\composition\genSpace}\biggl\{\frac{1}{\samplesizeNoise}\sum_{i=1}^{\samplesizeNoise}(\discriminator\composition\generator)(\noise_i)-\E_{\noise_i\sim\distriN}[\discriminator\composition\generator]\biggr\}}+3\epsilon\\
		&= 2\approxErr(\evaSpace,\disSpace,\domain)+\sup_{\discriminator\in\disSpace}\bigl\{|\E_{\distriT}[\discriminator]-\MOM_{\blockNum}[\discriminator]|\bigr\}+\inf_{\generator\in\genSpace}\sup_{\discriminator\in\disSpace}\biggl\{\MOM_{\blockNum}[\discriminator]-\frac{1}{\samplesizeNoise}\sum_{i=1}^{\samplesizeNoise}\discriminator(\generator(\noise_i))\biggr\}+{\distance_{\disSpace \circ \genSpace}(\widehat{\nu}_{m}, \nu)}+3\epsilon\\
		&= 2\approxErr(\evaSpace,\disSpace,\domain)+\sup_{\discriminator\in\disSpace}\bigl\{|\E_{\distriT}[\discriminator]-\MOM_{\blockNum}[\discriminator]|\bigr\}+\inf_{\generator\in\genSpace}\sup_{\discriminator\in\disSpace}\biggl\{\MOM_{\blockNum}[\discriminator]-\E_{\generator_{\#}\distriN}[\discriminator]+{\E_{\generator_{\#}\distriN}[\discriminator]-\frac{1}{\samplesizeNoise}\sum_{i=1}^{\samplesizeNoise}\discriminator(\generator(\noise_i))}\biggr\}\\
		&+\distance_{\disSpace \circ \genSpace}(\widehat{\nu}_{m}, \nu)+3\epsilon\\
		&\le 2\approxErr(\evaSpace,\disSpace,\domain)+\sup_{\discriminator\in\disSpace}\{|\E_{\distriT}[\discriminator]-\MOM_{\blockNum}[\discriminator]|\}+\inf_{\generator\in\genSpace}\sup_{\discriminator\in\disSpace}\{\MOM_{\blockNum}[\discriminator]-\E_{\generator_{\#}\distriN}[\discriminator]\}+2\distance_{\disSpace \circ \genSpace}(\widehat{\nu}_{m}, \nu)+3\epsilon
	\end{align*}
	where we define $\widehat{\nu}_{m}:={\samplesizeNoise}^{-1}\sum_{i=1}^\samplesizeNoise\delta_{\noise_i}$, and the last inequality is from
	\begin{align*}
		{\sup_{\discriminator\in\disSpace}\{\E_{\generator_{\#}\distriN}[\discriminator]-\frac{1}{\samplesizeNoise}\sum_{i=1}^{\samplesizeNoise}\discriminator(\generator(\noise_i))\}}
		&\le \sup_{\discriminator\in\disSpace}\biggl|\frac{1}{\samplesizeNoise}\sum_{i=1}^{\samplesizeNoise}\bigl((\discriminator\composition\generator)(\noise_i)-\E_{\generator_{\#}\distriN}[\discriminator]\bigr)\biggr|=\sup_{\discriminator\in\disSpace}\biggl|\frac{1}{\samplesizeNoise}\sum_{i=1}^{\samplesizeNoise}\bigl((\discriminator\composition\generator)(\noise_i)-\E_{\noise_i\sim\distriN}[(\discriminator\composition\generator)(\noise_i)]\bigr)\biggr|\\
		&\le {\sup_{\discriminator\composition\generator\in\disSpace\composition\genSpace}\biggl|\frac{1}{\samplesizeNoise}\sum_{i=1}^{\samplesizeNoise}\bigl((\discriminator\composition\generator)(\noise_i)-\E_{\noise_i\sim\distriN}[(\discriminator\composition\generator)(\noise_i)]\bigr)\biggr|=:\distance_{\disSpace \circ \genSpace}(\widehat{\nu}_{m}, \nu)},
	\end{align*}
	by the definition of IPM. 
	
	Finally, letting $\epsilon\rightarrow 0$ yields the desired result.
	
	\bigskip

	\noindent\textbf{Proof of Proposition~\ref{lem:our} }
	Recall that $K_S$ represents the number of sane block containing no outliers, and $\eta(\varepsilon)>0$ is a fraction function for sane block such that
	$K_S \ge \eta(\varepsilon) K$. For $e>0$, in fact, if
	\begin{center}
		$\sup_{f \in \mathcal{F}}\sum_{k \in [K_S]}1_{\{|(\E_{B_k}-\E)f|> e\}} \le K_S-K/2,$
		then $\sup_{f \in \mathcal{F}} \bigl|\MOM_\blockNum[f]-\E_{\distriT}[f]\bigr|\le e$.
	\end{center}
	\noindent The reason is that if at least $K/2$ sane block $\{B_k\}$ such that $|(\E_{B_k}-\E)f|
	\le e$ for all ${f \in \mathcal{F}}$, then one must have $\sup_{f \in \mathcal{F}} \bigl|\MOM_\blockNum[f]-\E_{\distriT}[f]\bigr|\le e$. Rigorously, we have
	$$
	\begin{aligned}
		\{\sup_{f \in \mathcal{F}} \bigl|\MOM_\blockNum[f]-\E_{\distriT}[f]\bigr|\le e\} & \supset\left\{\left|\left\{k \in [K_S]:|(\E_{B_k}-\E)f|\le e\right\}\right| \ge \frac{K}{2}, \forall~{f \in \mathcal{F}}\right\} \\
		& =\left\{\sum_{k \in [K_S]} 1_{\left\{|(\E_{B_k}-\E)f|> e\right\}} \le K_S-\frac{K}{2}, \forall~{f \in \mathcal{F}}\right\} \\
		&=\left\{\sup_{f \in \mathcal{F}}\sum_{k \in [K_S]} 1_{\left\{|(\E_{B_k}-\E)f|> e\right\}} \le K_S-\frac{K}{2}\right\}.
	\end{aligned}
	$$
	Then,
	\begin{align}\label{eq:supmom}
		P\left\{\sup_{f \in \mathcal{F}} \bigl|\MOM_\blockNum[f]-\E_{\distriT}[f]\bigr|\le e\right\}\ge P\left\{\sup_{f \in \mathcal{F}}\sum_{k \in [K_S]} 1_{\left\{|(\E_{B_k}-\E)f|> e\right\}} \le K_S-\frac{K}{2}\right\}.
	\end{align}
	
	It remains to find a high probability upper bound for $\sup_{f \in \mathcal{F}}\sum_{k \in [K_S]} 1_{\left\{|(\E_{B_k}-\E)f|>e\right\}}$. To this end, suppose that $\phi(\cdot)$ is 1-Lipschitz function such that
	$1_{\{x\ge 2\}}(x) \le \phi(x)\le 1_{\{x\ge 1\}}(x).$ Therefore, one has
	\begin{align}\label{eq:i1i2}
		&~~~~\sup_{f \in \mathcal{F}}\sum_{k \in [K_S]} 1_{\left\{\frac{2|(\E_{B_k}-\E)f|}{e}\ge 2\right\}}\\
		& \le \sup_{f \in \mathcal{F}}\sum_{k \in [K_S]} \phi\left(\frac{2|(\E_{B_k}-\E)f|}{e}
		\right)\nonumber\\
		& \le \sup_{f \in \mathcal{F}}\sum_{k \in [K_S]} \E \phi\left(\frac{2|(\E_{B_k}-\E)f|}{e}
		\right)+\sup_{f \in \mathcal{F}}\sum_{k \in [K_S]} \left\{\phi\left(\frac{2|(\E_{B_k}-\E)f|}{e}
		\right)-\E \phi\left(\frac{2|(\E_{B_k}-\E)f|}{e}
		\right)\right\}\nonumber\\
		& \le K_S\sup_{f \in \mathcal{F}}P{\left\{|(\E_{B_1}-\E)f|\ge e/2\right\}}+\sup_{f \in \mathcal{F}}\sum_{k \in [K_S]} \left\{\phi\left(\frac{2|(\E_{B_k}-\E)f|}{e}
		\right)-\E \phi\left(\frac{2|(\E_{B_k}-\E)f|}{e}
		\right)\right\}\nonumber\\
		&:=I_{1n}(e)+I_{2n}(e),
	\end{align}
	where the last inequality is by $\phi(x)\le 1_{\{x\ge 1\}}(x)$.

	The high probability upper bounds for $I_{2n}(e)$ is obtained by using McDiarmid's inequality (see Lemma 2.6 in \citet{zhang2020concentration}). Then, with probability at least $1-\exp(-2x^2/K_S)$, we have
	\begin{align}\label{eq:hpb}
		I_{2n}(e)&:=\sup_{f \in \mathcal{F}}\left\{\sum_{k \in [K_S]} \phi\left(\frac{2|(\E_{B_k}-\E)f|}{e}
		\right)-\E \phi\left(\frac{2|(\E_{B_k}-\E)f|}{e}
		\right)\right\}\nonumber\\
		&\le \E \sup_{f \in \mathcal{F}}\left\{\sum_{k \in [K_S]} \phi\left(\frac{2|(\E_{B_k}-\E)f|}{e}
		\right)-\E \phi\left(\frac{2|(\E_{B_k}-\E)f|}{e}
		\right)\right\}+x,~\forall~x \ge 0.
	\end{align}
	Next, we want to bound the expected term in the last expression by the symmetrization and contraction theorems. The symmetrization  theorem gives
	\begin{align*}
		\E \sup_{f \in \mathcal{F}}\left\{\sum_{k \in [K_S]} \phi\left(\frac{2|(\E_{B_k}-\E)f|}{e}
		\right)-\E \phi\left(\frac{2|(\E_{B_k}-\E)f|}{e}
		\right)\right\}&\le 2\E \sup_{f \in \mathcal{F}}\left\{\sum_{k \in [K_S]} \varRade_k\phi\left(\frac{2|(\E_{B_k}-\E)f|}{e}
		\right)\right\}\\
		[\text{By contraction theorem}]~& \le \frac{4}{e}\E \sup_{f \in \mathcal{F}}\left\{\sum_{k \in [K_S]} \varRade_k(\E_{B_k}-\E)f\right\}\\
		&=\frac{4K_S}{e}\E \sup_{f \in \mathcal{F}}\left\{ \varRade_1(\E_{B_1}-\E)f\right\}\\
		&\le \frac{4K_S}{e}[\E \sup_{f \in \mathcal{F}}\left\{ (\E_{B_1}-\E)f\right\}\frac{1}{2}+\E \sup_{f \in \mathcal{F}}\left\{ -(\E_{B_1}-\E)f\right\}\frac{1}{2}]\\
		[\text{By symmetrization theorem again}]~&\le \frac{8K_S\Rade_{n/K}(\mathcal{F}; \data)}{e},
	\end{align*}
	where the second inequality is from a sharper contraction theorem (see Proposition 4.3 and (4.9) in \cite{bach2021learning}).

	On the one hand, by \eqref{eq:i1i2} and \eqref{eq:hpb} we get
	\begin{align}\label{eq:supmom2}
		P\left(\sup_{f \in \mathcal{F}}\sum_{k \in [K_S]} 1_{\left\{{|(\E_{B_k}-\E)f|}\ge {e}\right\}}\le I_{1n}(e)+\frac{8K_S\Rade_{n/K}(\mathcal{F}; \data)}{e}+x\right)\ge 1-\exp(-2x^2/K_S).
	\end{align}
	
	On the other hand, the \eqref{eq:supmom} implies
	\begin{align}\label{eq:supmom1}
		P\left\{\sup_{f \in \mathcal{F}} \bigl|\MOM_\blockNum[f]-\E_{\distriT}[f]\bigr|\le e\right\}&\ge P\left\{\sup_{f \in \mathcal{F}}\sum_{k \in [K_S]} 1_{\left\{|(\E_{B_k}-\E)f|> e\right\}} \le K_S-\frac{K}{2}\right\}\nonumber\\
		& \ge P\left\{\sup_{f \in \mathcal{F}}\sum_{k \in [K_S]} 1_{\left\{|(\E_{B_k}-\E)f|> e\right\}} \le K_S\left(1-\frac{1}{2\eta(\varepsilon)}\right)\right\},
	\end{align}
	where the second inequality is due to $-\frac{K}{2} \ge -\frac{K_S}{2\eta(\varepsilon)}$.
	%\vee \sqrt{ \frac{16\sigma^{2} K^2}{n}}
	
	Put $x=\frac{K_S}{4}\left(1-\frac{1}{2\eta(\varepsilon)}\right)$ and suppose that $U_{1n}(e)$ satisfies
	$$I_{1n}(e):=K_S\sup_{f \in \mathcal{F}}P{\left\{|(\E_{B_1}-\E)f|\ge e/2\right\}}\le U_{1n}(e).$$
	
	Observe that if $e \ge \sqrt{\frac{4e^2U_{1n}(e) }{1-(2\eta(\varepsilon))^{-1}}}$ and thus $I_{1n}(e) \le U_{1n}(e) \le \frac{K_S}{4}\left(1-\frac{1}{2\eta(\varepsilon)}\right)$. Therefore, in \eqref{eq:supmom2}, one has
	\begin{align}\label{eq:supmom33}
		I_{1n}(e)+\frac{8K_S\Rade_{n/K}(\mathcal{F}; \data)}{e}+x \le \frac{K_S}{2}\left(1-\frac{1}{2\eta(\varepsilon)}\right)+\left(\frac{K_S}{4}+\frac{K_S}{4}\right)\left(1-\frac{1}{2\eta(\varepsilon)}\right)= K_S\left(1-\frac{1}{2\eta(\varepsilon)}\right),
	\end{align}
	from $e \ge \frac{16\Rade_{n/K}(\mathcal{F}; \data) }{1-(2\eta(\varepsilon))^{-1}}$. Then it implies via \eqref{eq:supmom1} with $e \ge \frac{16\Rade_{n/K}(\mathcal{F}; \data) }{1-(2\eta(\varepsilon))^{-1}} \vee \sqrt{\frac{4e^2U_{1n}(e) }{1-(2\eta(\varepsilon))^{-1}}}$
	\begin{align}\label{eq:supmom3}
		&~~~~P\left\{\sup_{f \in \mathcal{F}} \bigl|\MOM_\blockNum[f]-\E_{\distriT}[f]\bigr|\le \frac{16\Rade_{n/K}(\mathcal{F}; \data) }{1-(2\eta(\varepsilon))^{-1}} \vee \sqrt{\frac{4e^2U_{1n}(e) }{1-(2\eta(\varepsilon))^{-1}}}\right\}\nonumber\\
		& \ge P\left\{\sup_{f \in \mathcal{F}}\sum_{k \in [K_S]} 1_{\left\{|(\E_{B_k}-\E)f|> e\right\}} \le K_S\left(1-\frac{1}{2\eta(\varepsilon)}\right)\right\}\nonumber\\
		[\text{By}~\eqref{eq:supmom33}]~& \ge P\left(\sup_{f \in \mathcal{F}}\sum_{k \in [K_S]} 1_{\left\{{|(\E_{B_k}-\E)f|}\ge {e}\right\}}\le \frac{8K_S\Rade_{n/K}(\mathcal{F}; \data)}{e}+I_{1n}(e)+x\right)\nonumber\\
		[\text{By}~\eqref{eq:supmom2}]~&\ge 1-\exp\left(-\frac{1}{8}\left(1-\frac{1}{2\eta(\varepsilon)}\right)^2 K_S\right)\ge 1-\exp\left(-\frac{1}{8}\eta(\varepsilon)\left(1-\frac{1}{2\eta(\varepsilon)}\right)^2 K\right),
	\end{align}
	where the second inequality is from $K_S \ge \eta(\varepsilon) K$.\\
	
	\bigskip

	\noindent\textbf{Proof of Proposition~\ref{prop:genApproxErr} }
	Using the Proposition \ref{lem:our}, we aim to prove that for $\epsilon:=2\Rade_{(1-\varepsilon) n}(\disSpace; \data)+\frac{16\Rade_{n/K}(\disSpace; \data) }{1-(2\eta(\varepsilon))^{-1}}\vee \sqrt{ \frac{[(16\varianceDis)\wedge(32B^2/{\rm{e}})] K^2}{[1-(2\eta(\varepsilon))^{-1}]n}}+\lipSupB\sqrt{\frac{\blockNum}{(1-\varepsilon)n}}$, there exists a $\generator\in\genSpace$, the following probability inequality holds
	\begin{equation}
		\PP\Bigl\{\inf_{\generator\in\genSpace}\sup_{\discriminator\in\disSpace}\bigl\{\MOM_\blockNum[\discriminator]-\E_{\generator_{\#}\distriN}[\discriminator]\bigr\}> \epsilon\Bigr\}<{\rm{e}}^{-\blockNum/32}+{\rm{e}}^{-\frac{K}{8\eta(\varepsilon)}\left(1-\frac{1}{2\eta(\varepsilon)}\right)^2}.
	\end{equation}
	By applying the triangle inequality, we have
	\begin{align}\label{eq:triangle}
		\PP\Bigl\{\inf_{\generator\in\genSpace}\sup_{\discriminator\in\disSpace}\bigl\{\MOM_\blockNum[\discriminator]-\E_{\generator_{\#}\distriN}[\discriminator]\bigr\}> \epsilon\Bigr\}	&\le \PP\Bigl\{\sup_{\discriminator\in\disSpace}\bigl\{\MOM_\blockNum[\discriminator]-\E_{\distriT}[\discriminator]\bigr\}> \epsilon_1\Bigr\}\nonumber\\
		&+\PP\Bigl\{\sup_{\discriminator\in\disSpace}\bigl\{\E_{\distriT}[\discriminator]-\E_{\distriTemp}[\discriminator]\bigr\}> \epsilon_2\Bigr\}+\PP\Bigl\{\inf_{\generator\in\genSpace}\sup_{\discriminator\in\disSpace}\bigl\{\E_{\distriTemp}[\discriminator]-\E_{\generator_{\#}\distriN}[\discriminator]\bigr\}> \epsilon_3\Bigr\},
	\end{align}
	where $\epsilon\ge\epsilon_1+\epsilon_2+\epsilon_3$ with $\epsilon_1,\epsilon_2,\epsilon_3\ge0$ being determined later, and $\distriTemp:=\frac{1}{(1-\varepsilon)\samplesize}\sum_{i \in I}\delta_{\data_i}$ denotes the empirical distribution of the inliers $\{\data_{i}\}_{i \in I}$.
	
	Now, we need to deal with the three items on the right hand side.
	The first item can be controlled by the concentration inequality of the MoM estimator.
	The second item can be controlled by the concentration inequality of suprema of empirical process.
	There exists $\generator\in\genSpace$ such that the second last item in \eqref{eq:triangle} can be proved to be 0 according to Lemma~3.2 of~\cite{Yang2021Capacity}. 
	
	\emph{Step~1:} We first deal with the first term. Let $\Rade_{n/K}(\disSpace;\data):=\E\sup_{\discriminator\in\disSpace}\Bigl|\frac{1}{n/K}\sum_{i \in B_1}\varRade_i\discriminator(\data_i)\Bigr|$
	be the Rademacher complexity of functional class $\disSpace$, and $\varRade_1,\dots,\varRade_n$ are independent Rademacher random variables, which are independent of $\{\data_i\}_{i\in B_1}$. By  Proposition \ref{lem:our}, we have for any $\blockNum\in\{1,\dots,\samplesize/2\}$,
	\begin{equation*}\label{eq:Lerasle2019}
		P\left\{\sup_{\discriminator\in\disSpace}\bigl\{\MOM_\blockNum[\discriminator]-\E_{\distriT}[\discriminator]\bigr\}>\frac{16\Rade_{n/K}(\disSpace; \data) }{1-(2\eta(\varepsilon))^{-1}} \vee \sqrt{\frac{4e^2U_{1n}(e) }{1-(2\eta(\varepsilon))^{-1}}}\right\}\le {\rm{e}}^{-\frac{K}{8\eta(\varepsilon)}\left(1-\frac{1}{2\eta(\varepsilon)}\right)^2},
	\end{equation*}
	where $U_{1n}(e)$ satisfies
	$K_S\sup_{\discriminator\in\disSpace}P{\left\{|(\E_{B_1}-\E)[d]|\ge e/2\right\}}\le U_{1n}(e)$ for $e \ge \frac{16\Rade_{n/K}(\disSpace; \data) }{1-(2\eta(\varepsilon))^{-1}} \vee \sqrt{\frac{4e^2U_{1n}(e) }{1-(2\eta(\varepsilon))^{-1}}}> 0$.
	
	The Chebyshev's inequality implies $I_{1n}(e) \le K_S \cdot \frac{{\mathrm{Var}[\E_{B_1}f]}}{(e/2)^2}=\frac{4\sigma^2K_SK}{e^2 n}\le \frac{4\sigma^2K^2}{e^2 n}=:U_{1n}(e)$. We choose $e \ge \frac{16\Rade_{n/K}(\disSpace; \data) }{1-(2\eta(\varepsilon))^{-1}} \vee \sqrt{\frac{16\sigma^2K^2}{((1-(2\eta(\varepsilon))^{-1})n}}$. Since $\sup_{\discriminator\in\disSpace}\normsup{\discriminator}\le\lipSupB$. If we use Hoeffding's inequality, 
	$$K_S\sup_{\discriminator\in\disSpace}P{\left\{|(\E_{B_1}-\E)[d]|\ge e/2\right\}}\le K{\rm{e}}^{-(n/K) e^2/(8B^2)}=:U_{1n}(e) $$
	for $e \ge \frac{16\Rade_{n/K}(\disSpace; \data) }{1-(2\eta(\varepsilon))^{-1}} \vee \sqrt{\frac{4K }{1-(2\eta(\varepsilon))^{-1}}e^2{\rm{e}}^{-(ne^2)/(8KB^2)}}$. Note that $\sup_{t>0}\frac{4K }{1-(2\eta(\varepsilon))^{-1}}t^2{\rm{e}}^{-(nt^2) /(8KB^2)}=\frac{32K^2B^2/({\rm{e}}n) }{1-(2\eta(\varepsilon))^{-1}}$ for $t=2B\sqrt{\frac{2K}{n}}$. We choose $e \ge \frac{16\Rade_{n/K}(\disSpace; \data) }{1-(2\eta(\varepsilon))^{-1}} \vee \sqrt{\frac{(32B^2/{\rm{e}})K^2}{((1-(2\eta(\varepsilon))^{-1})n}}$.
	
	Combining the two choices above, we define $\epsilon_1:=\frac{16\Rade_{n/K}(\disSpace; \data) }{1-(2\eta(\varepsilon))^{-1}}\vee \sqrt{ \frac{[(16\varianceDis)\wedge(32B^2/{\rm{e}})] K^2}{[1-(2\eta(\varepsilon))^{-1}]n}}$. Hence,
	\begin{equation}\label{eq:Lerasle2019}
		P\left\{\sup_{\discriminator\in\disSpace}\bigl\{\MOM_\blockNum[\discriminator]-\E_{\distriT}[\discriminator]\bigr\}>\frac{16\Rade_{n/K}(\disSpace; \data) }{1-(2\eta(\varepsilon))^{-1}} \vee \sqrt{ \frac{[(16\varianceDis)\wedge(32B^2/{\rm{e}})] K^2}{[1-(2\eta(\varepsilon))^{-1}]n}}\right\}\le {\rm{e}}^{-\frac{K}{8\eta(\varepsilon)}\left(1-\frac{1}{2\eta(\varepsilon)}\right)^2}.
	\end{equation}

	% Collecting all the pieces above yields
	% \begin{align*}
		%     \PP\Bigl\{\sup_{\discriminator\in\disSpace}\bigl\{\MOM_\blockNum[\discriminator]-\E_{\distriT}[\discriminator]\bigr\}> \epsilon_1\Bigr\}\le e^{-\blockNum/32},
		% \end{align*}
	% where $\epsilon_1:=\frac{{256r\sqrt L_1 {(aM)^{L_1}}}}{{\sqrt n }}{\E}\mathop {\max }\limits_{i \in [n]}{\left\| \data_i \right\|_2}\vee 4\sdDis\sqrt{\frac{2\blockNum}{\samplesize}}$.
	
	\emph{Step~2:} We estimate the second term by a concentration inequality of suprema of empirical process.
	Observe that
	\begin{align*}
		\sup_{\discriminator\in\disSpace}\bigl\{\E_{\distriT}[\discriminator]-\E_{\distriTemp}[\discriminator]\bigr\}\le\sup_{\discriminator\in\disSpace}|\E_{\distriTemp}[\discriminator]-\E_{\distriT}[\discriminator]|=\sup_{\discriminator\in\disSpace}\biggl|\frac{1}{(1-\varepsilon)\samplesize}\sum_{i \in I}\Bigl(\discriminator(\data_i)-\E_{\data_i\sim\distriT}[\discriminator(\data_i)]\Bigr)\biggr|.
	\end{align*}
	Denote $Z:=\sup_{\discriminator\in\disSpace}\Bigl|\frac{1}{(1-\varepsilon)\samplesize}\sum_{i \in I}\bigl(\discriminator(\data_i)-\E_{\data_i\sim\distriT}[\discriminator(\data_i)]\bigr)\Bigr|$.
	Since $\E\bigl(\discriminator(\data_i)-\E_{\data_i\sim\distriT}[\discriminator(\data_i)]\bigr)=0$ and $$\bigl|\discriminator(\data_i)-\E_{\data_i\sim\distriT}[\discriminator(\data_i)]\bigr|\le\bigl|\discriminator(\data_i)\bigr|+\E_{\data_i\sim\distriT}\bigl|\discriminator(\data_i)\bigr|\le 2\lipSupB$$ for all $i\in I$, by Massart's concentration inequality for suprema of the bounded empirical processes, we have for all $t>0$
	\begin{align*}
		\PP\Biggl\{Z\ge\E Z + 2\lipSupB\sqrt{\frac{8t}{(1-\varepsilon)n}}\Biggr\}\le {\rm{e}}^{-t}.
	\end{align*}
	In addition, according to symmetrization theorem,
	\begin{align*}
		\E Z&=\E \sup_{\discriminator\in\disSpace}\Bigl|\frac{1}{(1-\varepsilon)\samplesize}\sum_{i \in I}\bigl(\discriminator(\data_i)-\E_{\data_i\sim\distriT}[\discriminator(\data_i)]\bigr)\Bigr|\le 2\E\sup_{\discriminator\in\disSpace}\Bigl|\frac{1}{(1-\varepsilon)\samplesize}\sum_{i \in I}\varRade_i\discriminator(\data_i)\Bigr|=2\Rade_{{(1-\varepsilon) n}}(\disSpace; \data).
	\end{align*}
	% 	According to the result that we have proved in Step~1, we have
	% 	\begin{equation*}
		% 	    \E\sup_{\discriminator\in\disSpace}\Bigl|\frac{1}{\samplesize}\sum_{i=1}^\samplesize\varRade_i\discriminator(\data_i)\Bigr|\le\frac{{2r\sqrt L {(aM)^L}}}{{\sqrt n }}{\E}\mathop {\max }\limits_{i \in [n]}{\left\| \data_i \right\|_2}.
		% 	\end{equation*}
	Hence, choosing $t:=\blockNum/32$, we have
	\begin{align*}
		\PP\Biggl\{\sup_{\discriminator\in\disSpace}\bigl\{\E_{\distriT}[\discriminator]-\E_{\distriTemp}[\discriminator]\bigr\}> 2\Rade_{(1-\varepsilon) n}(\disSpace; \data)+\lipSupB\sqrt{\frac{\blockNum}{(1-\varepsilon)n}}\Biggr\}\le {\rm{e}}^{-\blockNum/32}.
	\end{align*}
	and let $\epsilon_2:=2\Rade_{(1-\varepsilon) n}(\disSpace; \data)+\lipSupB\sqrt{\frac{\blockNum}{(1-\varepsilon)n}}$.

	\emph{Step~3:} We estimate the last item in \eqref{eq:triangle}. Suppose that  $\genSpace=\nn(\genWidth,\genDepth)$ satisfies $\genWidth\ge7\dimInput+1$, $\genDepth\ge2$ and
	\begin{center}
		$(1-\varepsilon)n\leq \frac{\genWidth-p-1}{2}\left\lfloor\frac{\genWidth-p-1}{6 p}\right\rfloor\left\lfloor\frac{\genDepth}{2}\right\rfloor+2 .$
	\end{center}
	
	For any $\disSpace\subseteq \operatorname{Lip}\left([0,1]^{p}, c\right)$, we have
	\begin{align*}
		\inf_{\generator\in\genSpace}\sup_{\discriminator\in\disSpace}\{\E_{\distriTemp}[\discriminator]-\E_{\generator_{\#}\distriN}[\discriminator]\}&=\inf _{g \in \mathcal{G}}\sup _{f \in \mathcal{F}}\int f(d \hat{\mu}_{n}-d g_{\#} \nu) \leq \inf _{g \in \mathcal{G}}c\sup _{\|f\|_{\rm Lip} \leq 1}\int f(d \hat{\mu}_{n}-d g_{\#} \nu)\\
		&= c \inf{}_{g \in \mathcal{G}} \mathcal{W}_{1}\left(\widehat{\mu}_{n}, g_{\#} \nu\right)=0,
	\end{align*}
	{$\text{since}~\mathcal{W}_{1}\left(\widehat{\mu}_{n}, {g_\epsilon}_{\#} \nu\right)\to 0~\text{for some}~g_\epsilon$} as $\epsilon \to 0$ by Lemma~3.2 of~Yang et al. (2022). Hence, for any $\epsilon_3\ge0$,
	\begin{equation*}
		\PP\Bigl\{\inf_{\generator\in\genSpace}\sup_{\discriminator\in\disSpace}\bigl\{\E_{\distriTemp}[\discriminator]-\E_{\generator_{\#}\distriN}[\discriminator]\bigr\}> \epsilon_3\Bigr\}=0.
	\end{equation*}
	Hence, one can choose $\epsilon_3:=0$.\\
	
	Finally, combining all the results above and the inequality $\epsilon\ge\epsilon_1+\epsilon_2+\epsilon_3$ yields
	\begin{align*}
		&\PP\Biggl\{\sup_{\discriminator\in\disSpace}\bigl\{\MOM_\blockNum[\discriminator]-\E_{\generator_{\#}\distriN}[\discriminator]\bigr\}> 2\Rade_{(1-\varepsilon) n}(\disSpace; \data)+\frac{16\Rade_{n/K}(\disSpace; \data) }{1-(2\eta(\varepsilon))^{-1}}\vee \sqrt{ \frac{[(16\varianceDis)\wedge(32B^2/{\rm{e}})] K^2}{[1-(2\eta(\varepsilon))^{-1}]n}}+\lipSupB\sqrt{\frac{\blockNum}{(1-\varepsilon)n}}\Biggr\}\\
		&\le {\rm{e}}^{-\blockNum/32}+{\rm{e}}^{-\frac{K}{8\eta(\varepsilon)}\left(1-\frac{1}{2\eta(\varepsilon)}\right)^2}.
	\end{align*}
	% 	where
	% 	$$\epsilon=\frac{{260r\sqrt L_1 {(aM)^{L_1}}}}{{\sqrt n }}{\E}\mathop {\max }\limits_{i \in [n]}{\left\| \data_i \right\|_2}+2(\lipSupB\vee 4\varianceDis\sqrt{2})\sqrt{\frac{\blockNum}{\samplesize}}.$$
	%The desired result is proved by $\PP(Y\le c)\ge 1-\PP(Y> c)$.
	
	\bigskip
	
	\noindent\textbf{Proof of Lemma~\ref{lem:RadComplexityNN} }
	We estimate the Rademacher complexity for DNNs functional class by the Lipschitz property of DNNs. Assume that the activation functions $\sigma_l: \mathbb{R}^{D_{l}} \rightarrow \mathbb{R}^{D_{l}}$ are $a$-Lipschitz w.r.t. Euclidean norm on their input and output spaces. Motivated by Lemma \ref{NN.xie} with function:
	\begin{align*}
		c_{\mathrm{Lip}}({\bm x}):=2a^{L} \sqrt{L}\|{\bm x}\|_{2} \max _{l \in\{0, \ldots, L\}} \prod_{j \in\{0, \ldots, L\}, j \neq l}\lambda_{\max }(W^{j}) \vee \lambda_{\max }(V^{j}),
	\end{align*}
	we suppose that the parameter set has $F$-norm ball with radius $r$ and  maximum spectral norm bounded by $M$,
	\begin{equation}\label{eq:FBALL}
		\bm\Theta:=\{\bm\theta:=( W_0, \ldots, W_{L} ):\|\bm\theta\|_{\mathrm{F}} \le r~\text{and}~\max_{0 \le l \le L}{\lambda _{\max }}({W_l}) \le {M}\} \subseteq \mathbb{R}^{S}.
	\end{equation}
	Let $\discriminator(\data)=f(\data;\bm\theta)$ as defined in \eqref{eq:NN}, and ${\discriminator\in\disSpace}$ is translated into $\bm\theta \in \bm\Theta$. Note that $f(\data_i;\bm\theta)=f(\data_i;\bm\theta)-f(\data_i;{\bf{0}})\le {c_{{\rm{Lip}}}}(\data_i){\left\| \bm\theta  \right\|_F}$ for $i=1,2,\cdots,n$, we have
	
	\allowdisplaybreaks
	\begin{align*}
		\E_\varRade\sup_{\discriminator\in\disSpace}\biggl|\frac{1}{\samplesize}\sum_{i=1}^\samplesize\varRade_i\discriminator(\data_i)\biggr|&=\E_\varRade\sup_{\bm\theta \in \bm\Theta}\biggl|\frac{1}{\samplesize}\sum_{i=1}^\samplesize\varRade_i[f(\data_i;\bm\theta)-f(\data_i;\bf{0})]\biggr|\\
		&\le\E_\varRade \mathop {\sup }\limits_{\bm\theta \in  \bm\Theta}\biggl| \frac{1}{n}\sum\limits_{i = 1}^n\varRade_i {c_{{\rm{Lip}}}}(\data_i){\left\| \bm\theta  \right\|_F}\biggr|\nonumber\\
		[\text{By}~\|\bm\theta\|_{\mathrm{F}}~\text{in}~\eqref{eq:FBALL}]~&\le \frac{r}{n}\E_\varRade\left| \sum\limits_{i = 1}^n \varRade_i{c_{{\rm{Lip}}}}(\data_i)\right| \le \frac{{r}}{n}\sqrt {\E_\varRade\left| \sum\limits_{i = 1}^n \varRade_i{c_{{\rm{Lip}}}}(\data_i)\right|^2}\nonumber\\
		&  = \frac{r}{n}\sqrt {\E_\varRade\left[ {\sum\limits_{i = 1}^n  c_{{\rm{Lip}}}^2(\data_i) + \sum\limits_{k \ne j} {{\varRade _k}} {\varRade _j}{c_{{\rm{Lip}}}}(\data_k){c_{{\rm{Lip}}}}(\data_j)} \right]}\le \frac{r}{{\sqrt n }}\sqrt{\mathop {\max }\limits_{k \in [n]} {c^2_{{\rm{Lip}}}}({\data_k}) } \\
		[\text{By}~\max_{0 \le l \le L}{\lambda _{\max }}({W_l}) \le {M}~\text{in}~\eqref{eq:FBALL}]~&\le \frac{{2r\sqrt L {(aM)^{L}}}}{{\sqrt n }}\mathop {\max }\limits_{i \in [n]}{\left\| \data_i \right\|_2},
	\end{align*}
	where the second inequality is from Ledoux-Talagrand contraction theorem (see Theorem 4.4 in \cite{ledoux1991probability}).
	
	Taking $\E_{\data}$ on the last expression, we have
	$\E\sup_{\discriminator\in\disSpace}\biggl|\frac{1}{\samplesize}\sum_{i=1}^\samplesize\varRade_i\discriminator(\data_i)\biggr|\le \frac{{2r\sqrt \Depth {(aM)^{\Depth}}}}{{\sqrt n }}{\E}\mathop {\max }\limits_{i \in [n]}{\left\| \data_i \right\|_2}.$

	\bigskip

	\begin{lemma}[Lipschitz property of neural networks, Proposition 6 in \cite{taheri2021statistical}]\label{NN.xie}
		For $l=0,1,\cdots,L$, assume that the activation functions $\sigma_l: \mathbb{R}^{N_{l}} \rightarrow \mathbb{R}^{N_{l}}$ is $a$-Lipschitz w.r.t. Euclidean norm on their input and output spaces. Then for every ${x} \in \mathbb{R}^{p}$ and parameters $\theta=\left(W^{0}, \ldots, W^{L}\right), \gamma=\left(V^{0}, \ldots, V^{L}\right)$ in \eqref{eq:NN},  it holds that
		$$\left|f(x;\theta ) - f(x;\gamma )\right| \leq c_{\mathrm{Lip}}({x})\|\theta-\gamma\|_{\mathrm{F}}$$
		with
		$c_{\mathrm{Lip}}({x}):=2a^{L} \sqrt{L}\|{x}\|_{2}\max _{l \in\{0, \ldots, L\}} \prod_{j \in\{0, \ldots, L\}, j \neq l}\sigma_{\max }(W^{j}) \vee \sigma_{\max }(V^{j}).$
	\end{lemma}
	
	\begin{lemma}[Lemma 2.7 in \cite{huang2021error}]\label{lem:huang}
		Assume $f \in \mathcal{H}^{b}([0,1]^{p})$ with ${b}=s+q, s \in \mathbb{N}_{0}$ and $q \in(0,1] .$ For any $N \geq 6$, $L \geq 2$, there exists
		\begin{center}
			$\phi \in \mathcal{N} \mathcal{N}\left(49(s+1)^{2} 3^{p} p^{s+1} N\left\lceil\log _{2} N\right\rceil, 15(s+1)^{2} L\left\lceil\log _{2} L\right\rceil+2 p\right)$
		\end{center}
		such that
		$\|\phi\|_{\infty} \leq 1$,
		$\|\phi\|_{\rm Lip}  \leq(s+1) p^{s+1 / 2} L(N L)^{\sigma(4 {b}-4) / p}(1260 N^{2} L^{2} 2^{L^{2}}+19 s 7^{s})$
		and
		\begin{center}
			$\|\phi-f\|_{L^{\infty}([0,1]^{p})} \leq 6(s+1)^{2} p^{(s+{b} / 2) \vee 1}\left\lfloor(N L)^{2 / p}\right\rfloor^{-{b}} .$
		\end{center}
	\end{lemma}

	%\end{appendices}
	
	%\section{Conclusion}\label{sec:conclusion}
	%In this paper, we proposed the MoM-GAN method to estimate an unknown probability measure $\distriT$ when data are contaminated. The generator and discriminator are both constructed by deep ReLU networks. We give the non-asymptotic IPM error bound for our estimator in Theorem~\ref{thm:main}, which depends on the sample size, the width and depth of neural network, and the input dimension.
	%To implement our method, we develop a gradient-based decent algorithm (see Algorithm~\ref{alg:the_alg}).
	%We study on two popular image datasets, MNIST and FashionMNIST, to exam our method and compare it with Wasserstein GAN method. The results show that the trained images by our method have higher resolution and lower FID than that of Wasserstein GAN.
	
	%For MoM-based method, the number of blocks $\blockNum$ is a crucial parameter. If $\blockNum$ is too large, the computational cost will be high and the upper bound of the evaluation error may not converge (see Theorem~\ref{thm:main}). If $\blockNum$ is too small, the robustness will be reduced and the probability bound will be poor (see Theorem~\ref{thm:main}). The cross-validation scheme may help to calibrate the parameter $\blockNum$, but its heavy computational consume is also a problem. So, to develop an efficient calibration scheme for the number of blocks should be a future work.
	
	%\section*{Acknowledgments}
	%
	%The authors would like to thank Emmanuel Cand\`es for several fruitful
	%discussions and the anonymous reviewers for very helpful suggestions.
	
	\bibliographystyle{apalike}
	\bibliography{references}

\end{document}

%% file: myMacros.tex
\usepackage{amsmath,amssymb,latexsym}
\usepackage{stmaryrd}
\usepackage{amsfonts,bbm,booktabs}
\usepackage{times,lastpage,bm,xspace}

\usepackage{theorem}
\newtheorem{remark}{Remark}
\newcommand{\squishlist}{
   \begin{list}{$\bullet$}
    { \setlength{\itemsep}{0pt}      \setlength{\parsep}{0pt}
      \setlength{\topsep}{0pt}       \setlength{\partopsep}{0pt}
      \setlength{\leftmargin}{1.5em} \setlength{\labelwidth}{1em}
      \setlength{\labelsep}{0.5em} } }

\newcommand{\squishlisttwo}{
   \begin{list}{$\bullet$}
    { \setlength{\itemsep}{0pt}    \setlength{\parsep}{0pt}
      \setlength{\topsep}{0pt}     \setlength{\partopsep}{0pt}
      \setlength{\leftmargin}{2em} \setlength{\labelwidth}{1.5em}
      \setlength{\labelsep}{0.5em} } }

\newcommand{\squishend}{
    \end{list}  }

\def\1{{\bf 1}}